\documentclass{article}
\usepackage{ijcai25}

\usepackage{times}
\usepackage{soul}
\usepackage{url}
\usepackage[hidelinks]{hyperref}
\usepackage[utf8]{inputenc}
\usepackage[small]{caption}
\usepackage{graphicx}
\usepackage{amsmath}
\usepackage{amsthm}
\usepackage{booktabs}
\usepackage{bbding}
\usepackage{amsmath,amssymb,amsfonts}
\usepackage{color}
\usepackage{relsize}
\usepackage{makecell}

\usepackage[switch]{lineno}
\usepackage[linesnumbered,ruled,vlined]{algorithm2e}

\def\eg{\emph{e.g., }}

\title{Multimodal Cancer Survival Analysis via Hypergraph Learning with Cross-Modality Rebalance}

\author{
Mingcheng Qu$^1$\and \hspace{-5pt}
Guang Yang$^1$\and \hspace{-5pt}
Donglin Di$^2$\and \hspace{-5pt}
Tonghua Su$^1$\and \hspace{-5pt}
Yue Gao$^2$\and \hspace{-5pt}
Yang Song$^3$\And \hspace{-5pt}
Lei Fan*$^{3}$
\\
\affiliations
$^1$Faculty of Computing, Harbin Institute of Technology\\
$^2$School of Software, Tsinghua University\\
$^3$School of Computer Science and Engineering, UNSW Sydney\\
\emails
*Corresponding author: lei.fan1@unsw.edu.au
}

\begin{document}

\maketitle

\begin{abstract}
    Multimodal pathology-genomic analysis has become increasingly prominent in cancer survival prediction. 
    However, existing studies mainly utilize multi-instance learning to aggregate patch-level features, neglecting the information loss of contextual and hierarchical details within pathology images.
    Furthermore, the disparity in data granularity and dimensionality between pathology and genomics leads to a significant modality imbalance. The high spatial resolution inherent in pathology data renders it a dominant role while overshadowing genomics in multimodal integration.
    In this paper, we propose a multimodal survival prediction framework that incorporates hypergraph learning to effectively capture both contextual and hierarchical details from pathology images. Moreover, it employs a modality rebalance mechanism and an interactive alignment fusion strategy to dynamically reweight the contributions of the two modalities, thereby mitigating the pathology-genomics imbalance. Quantitative and qualitative experiments are conducted on five TCGA datasets, demonstrating that our model outperforms advanced methods by over 3.4\% in C-Index performance\footnote{Code: \url{https://github.com/MCPathology/MRePath}}.
\end{abstract}

\section{Introduction}
Survival prediction, which focuses on forecasting events such as cancer progression and mortality in prognostic patients, is a critical area of research~\cite{nunes2024prognostic,fan2022cancer,fan2021learning}. Multimodal analysis has gained increasing prominence, exemplified by the integration of whole slide images (WSIs) and genomic profiles in clinical oncology~\cite{Mcat,MOTCat}.
The rationale behind this approach lies in the complementary strengths of each data modality: WSIs, regarded as the gold standard for cancer prognosis, provide cellular-level morphology and histopathological biomarkers~\cite{fan2022fast,tang2025prototype}, while genomic data yield essential molecular signatures and mutation profiles for accurate diagnosis~\cite{andre2022genomics,qu2024boundary}. 

In multimodal survival analysis, WSIs with giga-level resolution are typically segmented into multiple smaller patches to facilitate computationally efficient analysis~\cite{ilse2018attention}. 
Then, the multi-instance learning (MIL) technique~\cite{dietterich1997solving} is widely employed to aggregate patch-level features, operating under the assumption that a bag is labeled positive if it contains at least one positive instance; otherwise, it is considered negative.
Although MIL-based methods~\cite{hou2016patch,campanella2019clinical} excel in WSI classification tasks by effectively distinguishing tumor from non-tumor regions, they remain suboptimal for survival analysis. In many cases, they fail to capture the crucial contextual and hierarchical information required for precise prognostic assessment, leading to the inherent limitation of \textbf{information loss} in survival prediction. 
Specifically, contextual information~\cite{kapse2024si} includes spatial relationships and the micro-environment surrounding tumor cells, while hierarchical information~\cite{hou2022h} involves structural organization across various scales, \eg cellular, tissue, and organ levels. Both types of information are essential for accurate prognostic evaluations~\cite{zheng2018histopathological,jin2024hmil}.   

\begin{figure}[t]
    \centering
    \includegraphics[width=1\linewidth]{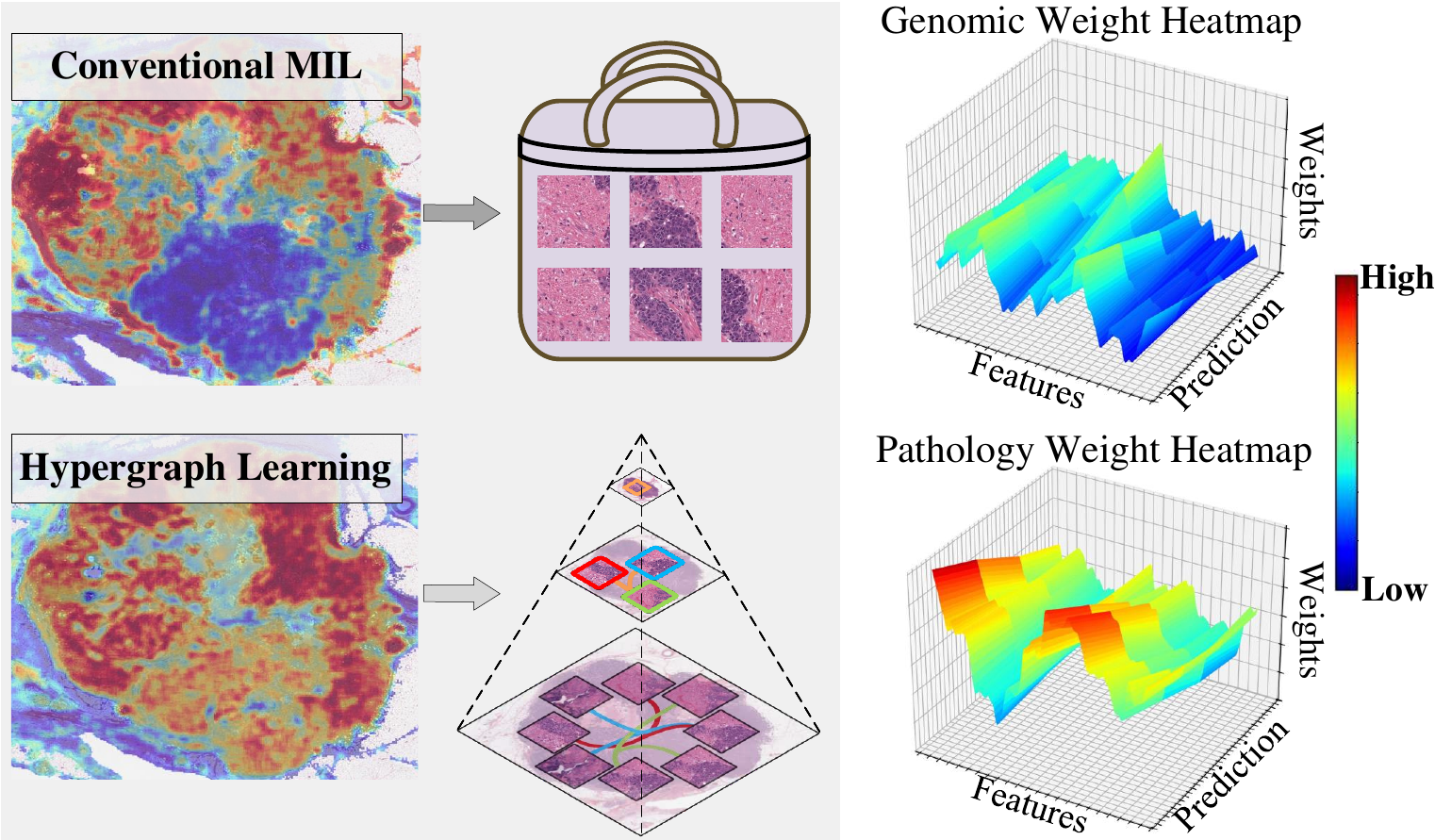}
    \caption{\textbf{Left:} Compared to MIL, hypergraph learning activates more patch regions and better captures contextual and hierarchical details. \textbf{Right:} Examples reveal the pathology-genomics imbalance, where pathology features dominate the overall survival prediction.}
    \label{fig:heatmap}
\end{figure}

On the other hand, integrating information from different modalities is a key research direction~\cite{lipkova2022artificial}. Existing methods often employ late fusion by simply combining two modalities~\cite{mobadersany2018predicting,cheerla2019deep,chen2020pathomic}, but they overlook the interconnections between genetic and pathologic data. Recent studies~\cite{SurvPath,zhang2024prototypical} explored early or mid fusion strategies, such as leveraging cross-attention mechanisms to better capture interactions across the modalities. However, these strategies still struggle to address issues related to \textbf{modality imbalance}. Specifically, a WSI can be represented through numerous patches, whereas only a few hundred genes have been identified for common cancers due to a higher signal-to-noise ratio
and lower information density~\cite{raser2005noise}. Consequently, this discrepancy can cause pathological data to dominate the fusion process and overshadow genomic data, 
especially when cross-attention mechanisms are utilized~\cite{Mcat,SurvPath}. As illustrated in Figure~\ref{fig:heatmap}, we conduct heuristic survival prediction experiments, where pathology modalities dominate the prediction task, highlighting the pathology-genomics imbalance.

In this paper, to address the information loss challenge, we aim to leverage graph learning as an alternative to MIL for aggregating patch-level features. Unlike existing graph-based approaches~\cite{shao2024tumor,chan2023histopathology} that reduce complex relationships to binary connections with limited expressiveness, we utilize hypergraphs~\cite{feng2019hypergraph} to model higher-order interactions among pathological features through hyperedges. This approach enables a more sophisticated representation of hierarchical and contextual details, enhancing the model's capability to capture intricate relationships. 
To tackle the challenge of pathology-genomics modality imbalance, we propose a dynamic weighting mechanism that adjusts the contribution of each modality based on its inherent reliability and synergy. By evaluating and aligning their interconnections, this approach ensures that each modality contributes optimally to the fusion process, effectively leveraging the strengths of both pathology and genomic data for more accurate survival predictions.

We propose a multimodal learning framework, MRePath (\textbf{M}ulti-Modal \textbf{Re}balance for \textbf{Path}ology-Genomic Survival Prediction), that leverages hypergraph learning and modality rebalance for survival analysis in WSIs. Specifically, in contrast to previous hypergraph studies~\cite{di2022big,li2023high,jing2025multi} that focus only on spatial relationships, we designate patches as nodes and construct hyperedges in both topological and feature spaces to capture spatial interactions and model structural hierarchies. We further introduce sheaf hypergraphs to promote information exchange between nodes and hyperedges, effectively preserving contextual details within individual patches and hierarchical relationships across the entire WSI.
For rebalancing pathology-genomics modalities, we employ a dynamic weighting mechanism that first assesses the mono-confidence of each modality's reliability and then calculates holo-confidence by evaluating their interactions. These assessments are integrated to determine the final weight, which rebalances the contribution of each modality.
Subsequently, an interactive alignment fusion is introduced, leveraging cross-attention to enable mutual guidance between modalities to produce the final hazard prediction. The contributions can be summarized as below:
\begin{itemize}
    \item A multimodal framework, MRePath, is proposed to address both the MIL-based information loss and the pathology–genomics modality imbalance challenges in survival analysis for WSIs.
    \item A hypergraph learning framework incorporating sheaf hypergraphs is constructed over both topological and feature spaces to capture contextual and hierarchical details, while enhancing the model's ability to differentiate between various types of information.  
    \item A modality rebalance method, consisting of a dynamic weighting mechanism and an interactive alignment fusion, is introduced to adjust the contributions of the two modalities for the final hazard prediction.
    \item Qualitative and quantitative experiments on five public datasets demonstrate the superiority of our model, achieving a 3.4$\%$ improvement over advanced methods.

\end{itemize}
\section{Related work}
\textbf{Survival analysis on WSIs.} 
Early studies typically rely on MIL~\cite{ilse2018attention} to aggregate patch-level features for representing WSIs~\cite{hou2016patch,campanella2019clinical}. Various methods have been used to extract global features, such as embeddings~\cite{yao2021deepprognosis}, attention weights~\cite{li2021dual}, and graph-based modeling~\cite{guan2022node,di2022big,di2022generating}. 
Recently, multimodal approaches integrating WSIs with genomic data for survival analysis have gained popularity~\cite{lipkova2022artificial}. Most research has focused on late fusion, such as vector concatenation~\cite{mobadersany2018predicting}, modality-level alignment~\cite{cheerla2019deep}, and bilinear pooling~\cite{chen2020pathomic}. 
Additionally, some studies have explored early or middle fusion approaches, leveraging cross-attention mechanisms for cross-modal interactions~\cite{Mcat,CMTA,SurvPath}.

Despite their promising performance, these approaches often overlook the modality imbalance. We propose a plug-and-play adjustment to evaluate each modality's reliability and connections, dynamically optimizing their contributions. This enhances multimodal fusion and improves performance.

\textbf{Graph-based pathology analysis.} 
Graph learning methods are widely adopted to capture intricate relationships in pathology-related tasks~\cite{guan2022node}. Early studies represented patches as graph nodes, constructing either adjacency-based~\cite{chen2021whole} or fully-connected graphs~\cite{adnan2020representation}. Recent studies have employed cellular graphs to reveal spatial relationships among specific biomarkers (\eg Ki67)~\cite{nakhli2023sparse} or cell types (\eg tumor and stromal cells)~\cite{shao2024tumor} within WSI. Additionally, some studies~\cite{di2022big,di2022generating} have utilized hypergraphs in WSIs for survival analysis, leveraging their representational power to capture complex interactions.

In our work, to tackle MIL-based information loss, we construct a hypergraph from both topological and feature spaces to capture contextual and hierarchical details. Additionally, we incorporate a sheaf hypergraph to adeptly differentiate and manage information from various types of hyperedges.

\begin{figure*}[ht]
    \centering
    \includegraphics[width=1\linewidth]{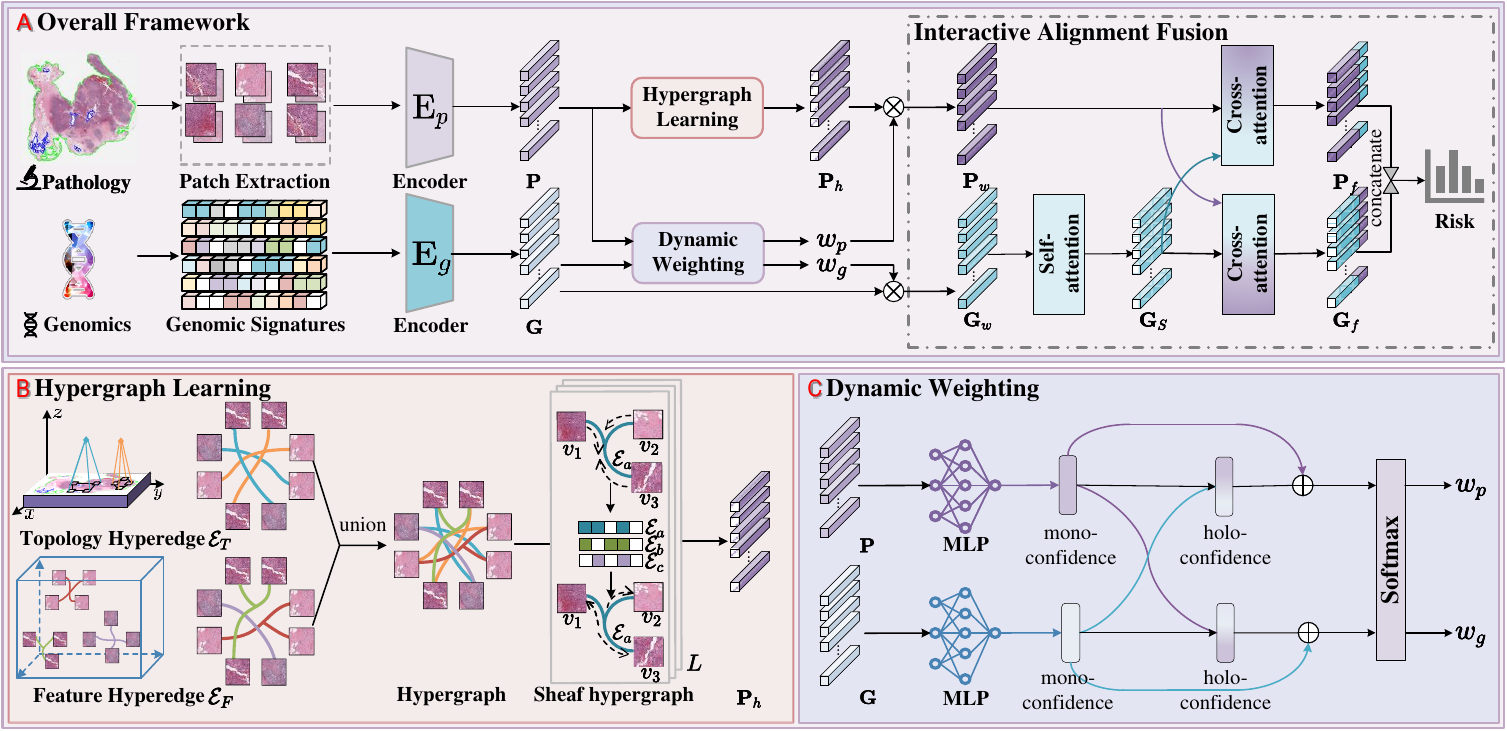}
    \caption{\textbf{Overview of MRePath.} \textbf{A,} MRePath consists of \textit{feature extraction} from pathology and genomics modalities, \textit{hypergraph learning} for capturing WSI representations, and \textit{modality rebalance} including dynamic weighting and interactive alignment fusion for recalibrating two modalities. \textbf{B,} Hypergraph learning involves constructing topological-based and feature-based hyperedges, and employing sheaf hypergraph to encourage local and global interactions among hyperedges, thereby capturing richer contextual and hierarchical details. \textbf{C,} Dynamic weighting computes mono-confidence and holo-confidence to produce modality weights for rebalancing modality contributions.}
    \label{fig:model}
\end{figure*}
\section{Method}
\subsection{Overview}
\textbf{Preliminary.} Given $\mathbb{X}=\{X_1, \dots, X_n\}$ is the cohort of $n$ subjects, each subject $X_i$ can be represented as a tuple $X_i=\left \{ \mathrm{H}_i,\mathrm{y}_i\right \}$. Here, $\mathrm{H}_i = \{\mathbf{P}_i, \mathbf{G}_i\}$ denotes a pair of pathology–genomics features, where $\mathbf{P}_i$ represents the WSI, and $\mathbf{G}_i$ represents the genomic profiles. Meanwhile, $\mathrm{y}_i = \{c_i, t_i\}$ represents the label of the $i$-th subject, comprising an event status $c_i \in \{0, 1\}$ ($c_i = 0$ indicates that the event has occurred) and the subject’s overall survival time $t_i$.
The goal of survival prediction is to estimate the hazard function $\phi_{h}(t)$, which predicts the instantaneous incidence rate of the interest event at a specific time point $t$. Instead of estimating a patient's survival time, we aim to train a model $\mathcal{F}$ to predict the probability that a patient's survival exceeds $t$ using the survival function $\phi_{s}(t)$. This process is supervised by the negative log-likelihood (NLL)~\cite{YAO2020101789}, defined as:
\begin{equation}
\begin{aligned}
    \mathcal{L}_{surv} =&-
\sum\nolimits_{i=1}^{n}\big\langle(1-c_i)\log{(\phi_h(t_i|\mathrm{H_i})}   
\\
&\hspace{-1.5em}+ c_i\log{\phi_s(t_i|\mathrm{H_i})} +(1-c_i)\log{\phi_{s}(t_i-1|\mathrm{H_i})}\big\rangle.
\end{aligned}
\end{equation}

\textbf{MRePath.}
It comprises three stages: feature extraction, hypergraph learning, and modality rebalance, as illustrated in Figure~\ref{fig:model}. 
Initially, features \( \mathbf{P} \) and \( \mathbf{G} \) are extracted from paired pathology and genomics data using different encoders. Hypergraph learning refines \( \mathbf{P} \) to produce \( \mathbf{P}_h \) with more contextual and hierarchical details. 
Modality rebalance includes a dynamic weighting module to obtain balanced features \( \mathbf{P}_w \) and \( \mathbf{G}_w \), and an interactive alignment fusion to produce integrated features \( \mathbf{P}_f \) and \( \mathbf{G}_f \) using cross-attention operations, predicting the final risk outcomes. 

\textbf{Pathology feature extraction}. Following previous studies~\cite{Mcat,MOTCat}, we partition each WSI into multiple $256\times256$ pixel patches at 20$\times$ magnification. 
A pretrained encoder model (\eg ResNet50) is used to extract $d$-dimensional features from these patches.
Each WSI is then represented as $\mathbf{P} \in \mathbb{R}^{N \times d} = \{p_1, p_2, \dots, p_N\}$, where $N$ is the number of patches, and the coordinates of the $ i $-th patch $p_i$ are $c_p^{(i)} = (x^{(i)}, y^{(i)})$.

\textbf{Genomic feature extraction}. 
To process genomic data, including RNA-seq, Copy Number Variation (CNV), and Simple Nucleotide Variation (SNV), which typically exhibit a high signal-to-noise ratio, a selection process is applied to enhance data quality~\cite{Mcat}.
The selected genes are categorized into six functional groups: Tumor Suppression, Oncogenesis, Protein Kinases, Cellular Differentiation, Transcription, and Cytokines and Growth.
These data are then embedded using a multilayer perceptron (MLP) to generate $ \mathbf{G} \in \mathbb{R}^{M \times d} = \{g_1, g_2, \dots, g_M\}$, where $ M $ represents the number of gene categories.

\subsection{Hypergraph Learning}

We aim to leverage hypergraph learning to capture contextual and hierarchical details from patch-level features in WSIs. Our approach consists of two key components: hypergraph construction, which encodes spatial and structural relationships within WSIs, and the sheaf hypergraph, which enhances representations through higher-order structures. Together, these components form a robust framework for representing spatial and structural patterns in WSIs, enabling more effective analysis of their complex relationships.

\textbf{Hypergraph construction}. A hypergraph $ \mathcal{G} = \{\mathcal{V}, \mathcal{E}\} $ is defined by a set of vertices $\mathcal{V}$ and a set of hyperedges $\mathcal{E}$. 
For a WSI $\mathbf{P}$, each patch is treated as a vertex, such that $\mathcal{V} = \{v_1, v_2, \dots, v_N\}$, where $N$ is the total number of patches. The feature $f_v^{(i)}$ of each vertex $v_i$ corresponds to the extracted features of its associated patch.

Hyperedges $\mathcal{E}$ are generated using two complementary methods: topological-based and feature-based approaches, capturing spatial and feature-level relationships, respectively. 
In the topological space, hyperedges are formed by grouping each patch with its neighboring patches based on the Euclidean distance in the spatial domain. Specifically, for a given patch $v_i$, its neighbors are determined as:
\begin{equation}
    \mathcal{N}_T(v_i) = \{v_j \mid \|c_p^{(i)} - c_p^{(j)}\|_2 \leq \delta \},
\end{equation}
where $c_p^{(i)}$ and $c_p^{(j)}$ denote the coordinates of patches $v_i$ and $v_j$, respectively, and $\delta$ is a distance threshold. This process results in the set of topological-based hyperedges $\mathcal{E}_T = \{\{{v_i, v_{j_1}, v_{j_2}, \dots \}} \mid \forall~v_j \in \mathcal{N}_T(v_i) \}$.

In the feature space, hyperedges are created based on the similarity between patch features. For a given patch $ v_i $, its feature-based neighbors are identified as:
\begin{equation}
    \mathcal{N}_F(v_i) = \{v_j \mid \text{sim}(f_v^{(i)}, f_v^{(j)}) \geq \gamma \},
\end{equation}
where $\text{sim}(\cdot, \cdot)$ is a similarity function, such as cosine similarity, and $\gamma$ is a similarity threshold. The values of \(\delta\) and \(\gamma\) are determined by the hyperedge construction threshold \(k\). Using these neighbors, the feature-based hyperedges are defined as $\mathcal{E}_F = \{\{v_i, v_{j_1}, v_{j_2}, \dots \} \mid \forall~ v_j \in \mathcal{N}_F(v_i) \}.$

The final hyperedge set $ \mathcal{E} $ is constructed by merging the topological-based and feature-based hyperedges, such that $\mathcal{E} = \mathcal{E}_T \cup \mathcal{E}_F$ where $\cup$ represents the take union.

\textbf{Sheaf hypergraph.} 
Given the constructed hypergraph $\mathcal{G}$, the pathology feature $ \mathbf{P} $ undergoes vanilla hypergraph convolution at the $ l $-th layer, expressed as:
\begin{equation}
\label{hg}
    \mathbf{P}^{(l+1)} = \sigma\left[(I_N - \Delta)\mathbf{P}^{(l)}\Theta^{(l)}\right],
\end{equation}
where $ \Delta $ represents the Laplacian operator, $ I_N $ is the identity matrix, $\Theta^{(l)}$ is a learnable weight matrix, and $\sigma$ is a nonlinear activation function. 
The sheaf hypergraph~\cite{duta2024sheaf} enhances this process by substituting the standard Laplacian operator \(\Delta\) in Eq.~\ref{hg} with the sheaf Laplacian \(\Delta_F\), allowing for data processing within a structured space over hyperedges, producing high-order features $\mathbf{P}_h = \mathbf{P}^{(L)}$ after $L$ layers. The sheaf Laplacian is defined as follows:
\begin{equation}
    \Delta_F = I_N - D_v^{-1/2} L_F D_v^{-1/2},
\end{equation}
where $ D_v $ is the degree matrix for vertices, and \( L_F \) is the sheaf Laplacian matrix, given by:
\begin{equation}
    L_F(v_i, v_j) = - \sum_{e; v_i, v_j \in e} D_e^{-1} F_{v_j \perp e}^\top F_{v_i \perp e}.
\end{equation}
Here, $F_{v \perp e}$ represents the linear maps that facilitate the flow of information from a vertex $ v $ to a hyperedge $ e $. It is achieved by averaging the features of the nodes within each hyperedge and then further aggregating these averaged features based on their association with each vertex.

The sheaf hypergraph creates an information flow space for nodes and hyperedges, enabling more expressive and structured dependencies for fine-grained information propagation. This approach effectively captures both local contextual features within individual patches and global hierarchical relationships across the entire WSI, resulting in a more robust representation. This framework is well-suited for survival prediction tasks requiring diverse information integration.

\subsection{Modality Rebalance}
We aim to achieve modality rebalance and interactive alignment fusion to recalibrate the contributions of each modality. This stage comprises two key modules: dynamic weighting, which calculates modality-specific weights to rebalance each modality, and interactive alignment fusion, which integrates the balanced features for final prediction.
\begin{table*}[h]
\caption{\textbf{Comparison of our MRePath and advanced methods on five datasets.} C-Index values (\%) are reported based on 5-fold cross-validation. ``p.” and ``g.” denote the pathology and genomics modalities, respectively, utilized by these methods.
Results of unimodal methods, SNN+CLAM, and Porpoise were cited from previous studies~\protect\cite{zhang2024prototypical}, while others were reproduced using their released codes.}
\label{table:comparison}
\centering
\resizebox{\textwidth}{!}{
\begin{tabular}{l|cc|lllll|c}
    \toprule
    Model & p. & g. & BLCA & BRCA & CO-READ & HNSC & STAD &Mean\\
    \cmidrule(lr){1-9}
    ABMIL~\cite{ilse2018attention} & \checkmark & &62.4 \textsmaller{$\pm$ 5.9} & 67.2 \textsmaller{$\pm$ 5.1} & 73.0 \textsmaller{$\pm$ 15.1} & 62.4 \textsmaller{$\pm$ 4.2} & 63.6 \textsmaller{$\pm$ 4.3} & 65.7  \\
    AMISL~\cite{YAO2020101789} & \checkmark &  & 62.7 \textsmaller{$\pm$ 3.2} & 68.1 \textsmaller{$\pm$ 3.6} & 71.0 \textsmaller{$\pm$ 9.1} & 60.7 \textsmaller{$\pm$ 4.8} & 55.3 \textsmaller{$\pm$ 1.2} & 63.6 \\
    TranMIL~\cite{shao2021transmil} & \checkmark &  & 61.7 \textsmaller{$\pm$ 4.5} & 66.3 \textsmaller{$\pm$ 5.3} & 74.7 \textsmaller{$\pm$ 15.1} & 61.9 \textsmaller{$\pm$ 6.2} & 66.0 \textsmaller{$\pm$ 7.2} & 66.1\\
    CLAM-SB~\cite{CLAM} & \checkmark &  & 64.3 \textsmaller{$\pm$ 4.4} & 67.5 \textsmaller{$\pm$ 7.4} & 71.7 \textsmaller{$\pm$ 17.2} & 63.0 \textsmaller{$\pm$ 4.8} & 61.6 \textsmaller{$\pm$ 7.8} & 65.6\\
    CLAM-MB~\cite{CLAM} & \checkmark &  & 62.3 \textsmaller{$\pm$ 4.5} & 69.6 \textsmaller{$\pm$ 9.8} & 72.1 \textsmaller{$\pm$ 15.9} & 64.8 \textsmaller{$\pm$ 5.0} & 62.0 \textsmaller{$\pm$ 3.4} & 66.2\\ 
    \cmidrule(lr){1-9}
    MLP~\cite{haykin1998neural} &  & \checkmark & 53.0 \textsmaller{$\pm$ 7.0}  & 62.2 \textsmaller{$\pm$ 7.9} & 71.2 \textsmaller{$\pm$ 11.4} & 52.0 \textsmaller{$\pm$ 6.4} & 49.7 \textsmaller{$\pm$ 3.1} & 57.6\\
    SNN~\cite{klambauer2017self} &  & \checkmark & 52.1 \textsmaller{$\pm$ 7.0} & 62.1 \textsmaller{$\pm$ 7.3} & 71.1 \textsmaller{$\pm$ 16.2} & 51.4 \textsmaller{$\pm$ 7.6} & 48.5 \textsmaller{$\pm$ 4.7} & 57.0\\
    SNNTrans~\cite{klambauer2017self} &  & \checkmark & 58.3 \textsmaller{$\pm$ 6.0} &  67.9 \textsmaller{$\pm$ 5.3}& 73.9 \textsmaller{$\pm$ 12.4} & 57.0 \textsmaller{$\pm$ 3.5} & 54.7 \textsmaller{$\pm$ 4.1} & 62.2\\
    \cmidrule(lr){1-9}
    SNN+CLAM& \checkmark & \checkmark & 62.5 \textsmaller{$\pm$ 6.0} & 69.9 \textsmaller{$\pm$ 6.4} & 71.6 \textsmaller{$\pm$ 16.0} & 63.8 \textsmaller{$\pm$ 6.6} & 62.9 \textsmaller{$\pm$ 6.5} & 66.1\\ 
    Porpoise~\cite{chen2022pan} &  \checkmark & \checkmark  & 61.7 \textsmaller{$\pm$ 5.6} & 66.8 \textsmaller{$\pm$ 7.0} & 73.8 \textsmaller{$\pm$ 15.1} & 61.4 \textsmaller{$\pm$ 5.8} & 66.0 \textsmaller{$\pm$ 10.6} & 65.9 \\ 
    MCAT~\cite{Mcat} & \checkmark & \checkmark & 64.0 \textsmaller{$\pm$ 7.6} & 68.5 \textsmaller{$\pm$ 10.9} & 72.4 \textsmaller{$\pm$ 13.7} &  56.4 \textsmaller{$\pm$ 8.4} & 62.5 \textsmaller{$\pm$ 11.8} & 64.8\\ 
    MOTCat~\cite{MOTCat} & \checkmark & \checkmark  & 65.9 \textsmaller{$\pm$ 6.9} & 72.7 \textsmaller{$\pm$ 2.7} & 74.2 \textsmaller{$\pm$ 12.4}  & 65.6 \textsmaller{$\pm$ 4.1} & 62.1 \textsmaller{$\pm$ 6.5} & 68.1\\ 
    CMTA~\cite{CMTA} & \checkmark & \checkmark & 67.0 \textsmaller{$\pm$ 3.0} & 69.1 \textsmaller{$\pm$ 3.7}& 70.4 \textsmaller{$\pm$ 11.7} & 56.2 \textsmaller{$\pm$ 8.6} & 59.2 \textsmaller{$\pm$ 1.4} &64.4 \\ 
    SurvPath~\cite{SurvPath} & \checkmark & \checkmark & 63.5 \textsmaller{$\pm$ 2.6} & 67.9 \textsmaller{$\pm$ 7.7} & 73.1 \textsmaller{$\pm$ 12.4} & 61.7 \textsmaller{$\pm$ 5.8} & 62.0 \textsmaller{$\pm$ 4.4} & 65.6\\ 
    PIBD~\cite{zhang2024prototypical} & \checkmark & \checkmark & 65.1 \textsmaller{$\pm$ 9.2} & 71.2 \textsmaller{$\pm$ 4.8}& 78.6 \textsmaller{$\pm$ 13.4} & 60.7 \textsmaller{$\pm$ 5.9} & 66.8 \textsmaller{$\pm$ 5.5} & 68.5\\ 
    \cmidrule(lr){1-9}
    \textbf{MRePath (Ours)} & \checkmark & \checkmark & \textbf{70.5 \textsmaller{$\pm$ 4.1}}  & \textbf{72.9 \textsmaller{$\pm$ 1.9}} & \textbf{80.8 $\pm$ \textsmaller{5.8}} & \textbf{66.0 \textsmaller{$\pm$ 5.8}} & \textbf{67.5 \textsmaller{$\pm$ 3.3}} & \textbf{71.5}\\ 
    \bottomrule
\end{tabular}
}
\end{table*}

\textbf{Dynamic weighting.}~It dynamically regulates the contributions of pathologic and genomic data by computing the weights $ w_p $ and $ w_g $ for their respective representations $ \mathbf{P} $ and $ \mathbf{G} $. This process leverages two confidence measures: \textit{mono-confidence}, which evaluates the reliability of individual modalities, and \textit{holo-confidence}, which captures their interactions. Together, these measures ensure effective modality fusion for accurate predictions.

Mono-confidence assesses the reliability of each modality by estimating the probability of accurately identifying the true class label within the respective dataset, thereby reflecting the modality's confidence~\cite{corbiere2019addressing}. The mono-confidence scores for pathology and genomic modalities are computed as:
\begin{equation}
    w_p^{m} = \mathbf{P} \Phi_p, \quad w_g^{m} = \mathbf{G} \Phi_g,
\end{equation}
where $ \Phi_p $ and $ \Phi_g $ are learnable parameters implemented with MLPs. A higher mono-confidence value indicates greater reliability of the corresponding modality~\cite{PDF}.

Holo-confidence extends mono-confidence by incorporating cross-modal interactions, reflecting the overall coordination and complementarity between pathology and genomics. The holo-confidence for each modality is computed as:
\begin{equation}
    w_p^{h} = \frac{\log(w_p^{m})}{\log(w_p^{m} \cdot w_g^{m})}, \quad
w_g^{h} = \frac{\log(w_g^{m})}{\log(w_p^{m} \cdot w_g^{m})}.
\end{equation}
These measures quantify how effectively each modality interacts with the other, providing a more holistic evaluation of their contributions. 

To obtain the final weights, mono-confidence and holo-confidence are combined through a linear operation followed by a softmax to convert probabilities, defined as:
\begin{equation}
w_p, w_g = \phi(w_p^{m} + w_p^{h}, w_g^{m} + w_g^{h})
\end{equation}
where $\phi$ denotes the softmax, $w_p$ and $w_g$ are the final weights for pathology and genomic features, respectively. These weights are used to adjust the pathology and genomic features, resulting in weighted representations $\mathbf{P}_w=w_p\mathbf{P}_h$ and $\mathbf{G}_w=w_g\mathbf{G}$.
This dynamic weighting mechanism enables the model to adaptively balance the contributions of the two modalities, leveraging both their individual reliability and their interactions to achieve optimal fusion.

\textbf{Interactive alignment fusion.}
The integration of pathology and genomic features is further enhanced by an interactive alignment strategy, which captures the internal relationships between the two modalities and facilitates their effective fusion. This process involves modality-specific co-attention mechanisms to refine feature representations.

To enhance the selection of pathological features based on genomic information, a gene-guided co-attention layer $ \mathcal{A}_{G \rightarrow P} $ is employed. This mechanism results in the $\mathbf{P}_{C}$, as:
\begin{equation}
    \mathbf{P}_{C} = \mathcal{A}_{G \rightarrow P} = \phi\left(\frac{w_p^q \mathbf{P}_w (w_p^k \mathbf{G}_f)^T}{\sqrt{d}}\right) w_p^v \mathbf{G}_f,
\end{equation}
where $\phi(\cdot)$ refers to the softmax function. $ w_p^k\mathbf{G}_f $, $ w_p^q \mathbf{P}_w $, and $ w_p^v\mathbf{G}_f $ are keys, queries, and values, respectively. This co-attention mechanism aligns genomic features $\mathbf{G}_f$ with pathology features $\mathbf{P}_w$, highlighting the most relevant pathological features. $ \mathbf{P}_{C} $ is then combined with $\mathbf{P}_w $ via a residual connection to produce the fused pathological features $ \mathbf{P}_f $.

For genomic features, a self-attention layer is first applied to capture intra-modal relationships, producing $\mathbf{G}_{S}$. Subsequently, a pathology-guided co-attention layer $ \mathcal{A}_{P \rightarrow G} $ is employed to enhance genomic feature selection based on pathological information. The co-attention $ \mathbf{G}_{C} $ are computed as:
\begin{equation}
    \begin{aligned}
        \mathbf{G}_{C} = \mathcal{A}_{P \rightarrow G} =\phi\left(\frac{w_g^q \mathbf{G}_{S}  (w_g^k\mathbf{P}_w)^T}{\sqrt{d}}\right) w_g^v\mathbf{P}_w,
    \end{aligned}
\end{equation}
where $ w_g^k\mathbf{P}_w $, $ w_g^q\mathbf{G}_{S} $, and $ w_g^v\mathbf{P}_w $ are keys, queries, and values in the co-attention process. The co-attended genomic features $\mathbf{G}_{C}$ are then combined with $ \mathbf{G}_{S} $ via residual connections to produce the refined genomic features $ \mathbf{G}_f $.

The interactive alignment strategy ensures that both modalities contribute complementary information while maintaining internal consistency. The fused pathology features $ \mathbf{P}_f $ and refined genomic features $ \mathbf{G}_f $ are enriched with modality-specific and cross-modality interactions, offering a robust and comprehensive representation for survival prediction.

\section{Experiments}

\subsection{Datasets and Settings}
\textbf{Datasets.} We followed previous studies~\cite{SurvPath,zhang2024prototypical} and selected five datasets from The Cancer Genome Atlas (TCGA) to evaluate the performance of our model. The datasets include: Bladder Urothelial Carcinoma (BLCA) (n=384), Breast Invasive Carcinoma (BRCA) (n=968), Colon and Rectum Adenocarcinoma (CO-READ) (n=298), Head and Neck Squamous Cell Carcinoma (HNSC) (n=392), and Stomach Adenocarcinoma (STAD) (n=317). The detailed distribution of survival times within these datasets is provided in the \textit{supplementary materials}.

\textbf{Evaluation metric.} We used the Concordance Index (C-Index) to assess the predictive accuracy of our model. For each cancer type, we conducted 5-fold cross-validation, splitting the data into training and validation sets with a 4:1 ratio. Results are reported as the mean C-Index $\pm$ standard deviation (STD) across the five datasets.

\textbf{Implementation details.} To ensure a fair comparison, we adopted similar settings as previous studies~\cite{Mcat,SurvPath,zhang2024prototypical}, using identical dataset splits and employing the Adam optimizer with a learning rate of $1 \times 10^{-4}$, a weight decay of $1 \times 10^{-5}$, and 30 training epochs.

\begin{table}[t]
\caption{\textbf{Ablation study on hypergraph learning.} Performance of various graph structures and hyperedge types, including topological-based hyperedges $\mathcal{E}_T$ and feature-based hyperedges $\mathcal{E}_F$, is reported in C-Index (\%).}

\label{table:ablation-graph}
\centering
\resizebox{\linewidth}{!}{
\begin{tabular}{c|c|lllll|c}
    \toprule
   GNN & Hyperedges & BLCA & BRCA & CO-READ & HNSC & STAD &Mean\\
    \cmidrule(lr){1-8}
    / & /  & \fontsize{14}{16}\selectfont 62.5\fontsize{10}{12}\selectfont$\pm$ 3.5 & \fontsize{14}{16}\selectfont 71.5\fontsize{10}{12}\selectfont$\pm$ 4.2 & \fontsize{14}{16}\selectfont 69.9\fontsize{10}{12}\selectfont$\pm$ 9.0 & \fontsize{14}{16}\selectfont 63.9\fontsize{10}{12}\selectfont$\pm$ 3.9 & \fontsize{14}{16}\selectfont 61.6\fontsize{10}{12}\selectfont$\pm$ 5.1  & \fontsize{14}{16}\selectfont 65.9\\
    GAT& / & \fontsize{14}{16}\selectfont 69.5\fontsize{10}{12}\selectfont$\pm$ 6.3 & \fontsize{14}{16}\selectfont 67.7\fontsize{10}{12}\selectfont$\pm$ 1.8 & \fontsize{14}{16}\selectfont 71.0\fontsize{10}{12}\selectfont$\pm$ 5.9 & \fontsize{14}{16}\selectfont 62.6\fontsize{10}{12}\selectfont$\pm$ 5.4 & \fontsize{14}{16}\selectfont 64.5\fontsize{10}{12}\selectfont$\pm$ 5.7 & \fontsize{14}{16}\selectfont 67.1\\
    GCN& / & \fontsize{14}{16}\selectfont 69.0\fontsize{10}{12}\selectfont$\pm$ 3.6 & \fontsize{14}{16}\selectfont 69.5\fontsize{10}{12}\selectfont$\pm$ 3.3 & \fontsize{14}{16}\selectfont 72.6\fontsize{10}{12}\selectfont$\pm$ 9.8 & \fontsize{14}{16}\selectfont 60.1\fontsize{10}{12}\selectfont$\pm$ 2.1 & \fontsize{14}{16}\selectfont 66.9\fontsize{10}{12}\selectfont$\pm$ 6.6  & \fontsize{14}{16}\selectfont 67.6\\
    HGNN& $\mathcal{E}_T$+$\mathcal{E}_F$ &\fontsize{14}{16}\selectfont 69.2\fontsize{10}{12}\selectfont$\pm$ 2.2 & \fontsize{14}{16}\selectfont 72.4\fontsize{10}{12}\selectfont$\pm$ 5.7 &\fontsize{14}{16}\selectfont 80.0\fontsize{10}{12}\selectfont$\pm$ 6.3 & \fontsize{14}{16}\selectfont 64.9\fontsize{10}{12}\selectfont$\pm$ 2.5 & \fontsize{14}{16}\selectfont 66.1\fontsize{10}{12}\selectfont$\pm$ 5.2  &\fontsize{14}{16}\selectfont 70.5\\
    \cmidrule(lr){1-8}
    SHGNN& $\mathcal{E}_T$ & \fontsize{14}{16}\selectfont 68.4\fontsize{10}{12}\selectfont$\pm$ 2.9 & \fontsize{14}{16}\selectfont 70.1\fontsize{10}{12}\selectfont$\pm$ 2.6 & \fontsize{14}{16}\selectfont 77.1\fontsize{10}{12}\selectfont$\pm$ 8.4 & \fontsize{14}{16}\selectfont 62.4\fontsize{10}{12}\selectfont$\pm$ 1.7 &  \fontsize{14}{16}\selectfont 65.3\fontsize{10}{12}\selectfont$\pm$ 5.3 & \fontsize{14}{16}\selectfont 68.7\\
    SHGNN& $\mathcal{E}_F$ & \fontsize{14}{16}\selectfont 69.3\fontsize{10}{12}\selectfont$\pm$ 2.2 & \fontsize{14}{16}\selectfont70.6\fontsize{10}{12}\selectfont$\pm$ 3.7 & \fontsize{14}{16}\selectfont 73.5\fontsize{10}{12}\selectfont$\pm$ 7.2 & \fontsize{14}{16}\selectfont 65.1\fontsize{10}{12}\selectfont$\pm$ 7.2 &  \fontsize{14}{16}\selectfont 65.3\fontsize{10}{12}\selectfont$\pm$ 3.4 & \fontsize{14}{16}\selectfont 68.8 \\
    \cmidrule(lr){1-8}
    SHGNN&$\mathcal{E}_T$+$\mathcal{E}_F$ & \fontsize{14}{16}\selectfont \textbf{70.5}\fontsize{10}{12}\selectfont$\pm$ 4.1 & \fontsize{14}{16}\selectfont \textbf{72.9}\fontsize{10}{12}\selectfont$\pm$ 1.9 & \fontsize{14}{16}\selectfont \textbf{80.8}\fontsize{10}{12}\selectfont $\pm$ 5.8 & \fontsize{14}{16}\selectfont \textbf{66.0}\fontsize{10}{12}\selectfont $\pm$ 5.8 & \fontsize{14}{16}\selectfont \textbf{67.5}\fontsize{10}{12}\selectfont$\pm$ 3.3 & \fontsize{14}{16}\selectfont\textbf{71.5}  \\
    \bottomrule
\end{tabular}
}
\end{table}

\subsection{Comparisons and Results}
We classified the advanced methods into three distinct groups according to the modalities they used: pathology only (ABMIL, AMISL, TransMIL, and CLAM), genomic only (MLP, SNN, and SNNTrans), and multimodal approaches (Porpoise, MCAT, MOTCat, CMTA, SurvPath, and PIBD).

As shown in Table~\ref{table:comparison}, unimodal methods such as MLP and SNN achieved average C-Index scores of 57.6$\%$ and 57.0$\%$, respectively. These scores were significantly lower than those of multimodal methods such as MOTCat (68.1$\%$), SurvPath (65.6$\%$) and PIBD (68.5$\%$), underscoring the limitations of unimodal approaches in capturing complex biological information. Among multimodal strategies, early fusion methods (\eg MCAT, MOTCat, CMTA, SurPath, PIBD, and our approach) generally outperformed late fusion methods including SNN + CLAM (66.1$\%$) and Porpoise (65.9$\%$). This highlights the importance of effectively capturing interactions between modalities.
Our method consistently outperformed both unimodal and other multimodal approaches across all cancer datasets, achieving the highest overall C-Index of 71.5$\%$ and an improvement of 3$\%$ over the second-highest method PIBD. These results demonstrate the superior integration of pathology and genomics in our framework.

\subsection{Ablation Studies}

\textbf{Hypergraph learning.}
We evaluated various graph architectures and hypergraph configurations, as summarized in Table~\ref{table:ablation-graph}. The baseline model with only MLP for feature aggregation achieved a C-Index of 65.9$\%$. By employing GNN structures, GAT and GCN enhanced the C-Index to 67.1$\%$ and 67.6$\%$, respectively, showing the effectiveness of graph structure in capturing relational interactions. Leveraging HGNN further elevated the mean C-Index to 70.5$\%$, highlighting the utility of hyperedges in capturing high-order relationships compared to standard graph structures. 

Incorporating a sheaf hypergraph with only topological-based hyperedges $\mathcal{E}_T$ or feature-based hyperedges $\mathcal{E}_F$ yielded scores of 68.7$\%$ and 68.8$\%$, respectively, while combining both hyperedges achieved a superior C-Index of 71.5$\%$. This indicates that integrating topological and feature-based connections significantly enhances the feature representations of WSIs, and that sheaf hypergraphs encourage local and global interactions to capture richer contextual and hierarchical details. These results highlight the efficacy of hypergraphs in capturing intricate relationships within the pathology data.

\begin{figure}[!t]
    \centering
    \includegraphics[width=1\linewidth]{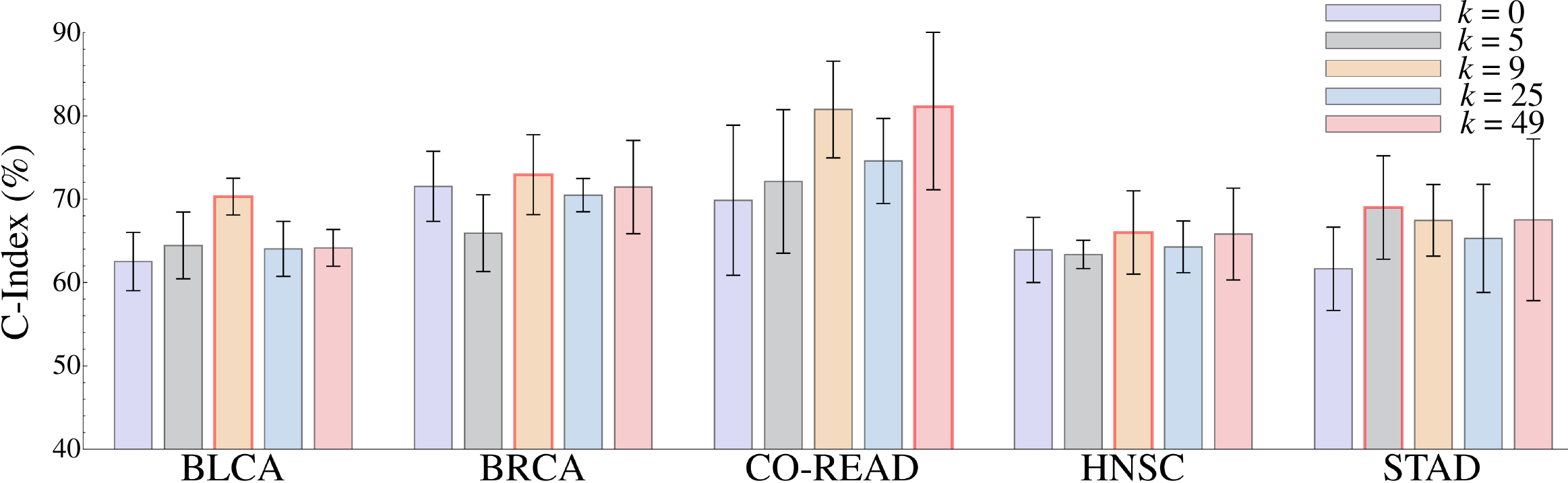}
    \caption{\textbf{Ablation study on hyperedge construction threshold $k$.} Performance of various similarity thresholds ($k$ = 0, 5, 9, 25, 49) is reported in C-Index ($\%$). }
    \label{fig:knn}
\end{figure}

\textbf{Hyperedge construction threshold.}
We evaluated various hyperedge construction threshold $k$, as illustrated in Figure~\ref{fig:knn}. The baseline model without hyperedges ($k$~=~0) showed moderate performance 
 across all five datasets. Increasing \( k \) enhanced performance, with the best results achieved at $k$~=~9. However, further increasing \( k \) led to a decline in performance and more computational costs. Although there was a slight improvement in CO-READ at $k$~=~49, the other datasets failed to regain their peak performance.
 The threshold \(k\) determines the range of node connections, balancing local and global information. Proper tuning of \(k\) enhances model performance by capturing meaningful relationships while avoiding information homogenization ($k$~=~49) or oversimplification ($k$~=~0).

\begin{figure*}[thbp]
    \centering
    \vspace{-10pt} 
    \includegraphics[width=1\linewidth]{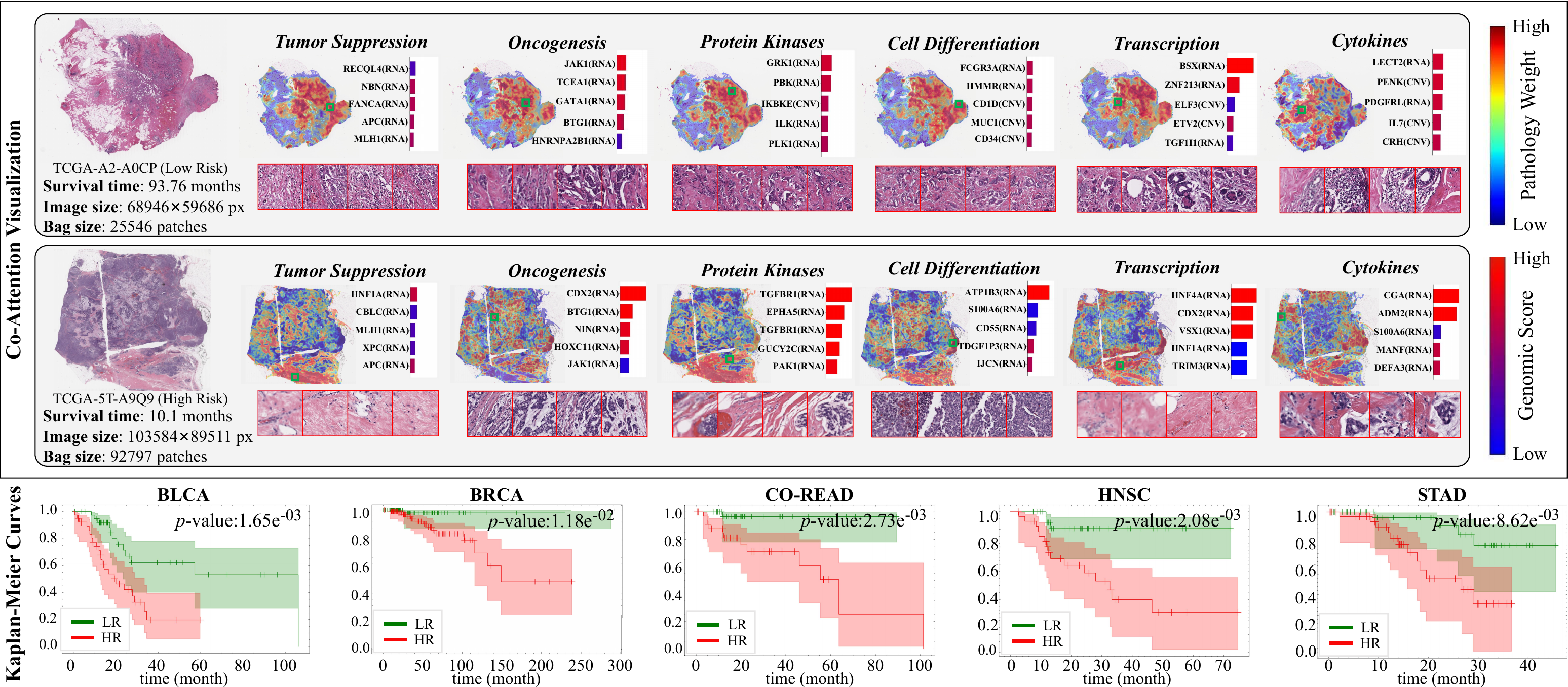}
    \caption{\textbf{Visualization for low and high-risk cases in the BRCA (Top):} The heatmaps are generated using cross-attention scores, with red and blue for high and low scores. The top five most influential genes are also highlighted in red for high and blue for low. \textbf{Kaplan-Meier curves (Bottom):} The high-risk and low-risk groups are determined based on the median predicted risk scores of our model. Our model achieved \textit{p}-values less than 0.05 across all five datasets, demonstrating its excellent stratification capability.}
    \label{fig:visualization}
\end{figure*}

\textbf{Modality rebalance.}
We evaluated the modality weighting and Interactive Alignment Fusion (IFA), as shown in Table~\ref{table:ablation-fusion}. We first manually assigned equal weights to pathology and genomics ($w_p$$=$$0.5,w_g$$=$$0.5$) with IFA produced a mean C-Index of 67.8$\%$. By using different weights to favor pathology ($w_p$$=$$0.7,w_g$$=$$0.3$), the mean C-Index increased to 69.3$\%$, which can be attributed to the effective extraction of pathology features through the sheaf hypergraph. However, it also exhibited higher instability, with the largest standard deviation observed in the BRCA, CO-READ, HNSC, and STAD datasets. Weighting the higher value to genomics ($w_p$$=$$0.3$, $w_g$$=$$0.7$) yielded a slightly higher score of 69.5$\%$, verifying the effectiveness of modality rebalance in modality fusion. We observed that introducing a dynamic weighting mechanism with IFA showed the best performance of 71.5$\%$, demonstrating the advantages of dynamically recalibrating each modality's contributions. 

When using dynamic weighting, models with different co-attention layers produced varied outcomes. Specifically, pathology-guided and gene-guided co-attention (P.G+G.P) achieved a mean C-Index of 66.9$\%$, while self-attention for genomics combined with either P.G or G.P yielded scores of 68.9$\%$ and 68.5$\%$, respectively. Compared to these settings, using IFA resulted in an improvement of 2.6$\%$ in the mean C-Index, highlighting the importance of capturing both internal interactions within each modality and their mutual influence during the modality fusion process.

\begin{table}[t]
\caption{\textbf{Ablation study on modality rebalance.} Performance of various modality weights and fusion strategies is reported in C-Index ($\%$). ``IFA." denotes the proposed Interactive Alignment Fusion. ``P.G", ``G.P", and ``SA." denote pathology-guided and gene-guided co-attention layers and self-attention for genomics respectively.}
\label{table:ablation-fusion}
\centering
\resizebox{\linewidth}{!}{

\begin{tabular}{c|c|lllll|c}
    \toprule
    Modality weighting & Fusion & BLCA & BRCA & CO-READ & HNSC & STAD &Mean\\
    \cmidrule(lr){1-8}
     $w_p = 0.5, w_g= 0.5$ & IFA. & \fontsize{14}{16}\selectfont 69.3 \fontsize{10}{12}\selectfont$\pm$ 3.0 & \fontsize{14}{16}\selectfont 67.2 \fontsize{10}{12}\selectfont$\pm$ 3.1 & \fontsize{14}{16}\selectfont 70.8 \fontsize{10}{12}\selectfont$\pm$ 8.4 & \fontsize{14}{16}\selectfont 65.1 \fontsize{10}{12}\selectfont$\pm$ 4.1 & \fontsize{14}{16}\selectfont 66.5 \fontsize{10}{12}\selectfont$\pm$ 5.3  & \fontsize{14}{16}\selectfont67.8\\

     $w_p = 0.7, w_g = 0.3$ & IFA. & \fontsize{14}{16}\selectfont 68.3 \fontsize{10}{12}\selectfont$\pm$ 2.5 & \fontsize{14}{16}\selectfont68.9 \fontsize{10}{12}\selectfont$\pm$ 5.5 & \fontsize{14}{16}\selectfont76.2 \fontsize{10}{12}\selectfont$\pm$ 11.3 & \fontsize{14}{16}\selectfont65.6 \fontsize{10}{12}\selectfont$\pm$ 5.5 & \fontsize{14}{16}\selectfont67.3 \fontsize{10}{12}\selectfont$\pm$ 8.0  & \fontsize{14}{16}\selectfont69.3\\

    $w_p=0.3,w_g=0.7$ & IFA.  & \fontsize{14}{16}\selectfont68.6 \fontsize{10}{12}\selectfont$\pm$ 3.5 & \fontsize{14}{16}\selectfont71.0 \fontsize{10}{12}\selectfont$\pm$ 4.6 & \fontsize{14}{16}\selectfont75.7 \fontsize{10}{12}\selectfont$\pm$ 5.7 & \fontsize{14}{16}\selectfont64.1 \fontsize{10}{12}\selectfont$\pm$ 4.1 & \fontsize{14}{16}\selectfont68.0 \fontsize{10}{12}\selectfont$\pm$ 4.7  & \fontsize{14}{16}\selectfont69.5\\
    \cmidrule(lr){1-8}

    Dynamic weighting & P.G+G.P  & \fontsize{14}{16}\selectfont 67.6 \fontsize{10}{12}\selectfont$\pm$ 3.6 & \fontsize{14}{16}\selectfont 69.1\fontsize{10}{12}\selectfont$\pm$ 2.3 & \fontsize{14}{16}\selectfont 71.1 \fontsize{10}{12}\selectfont$\pm$ 10.8 & \fontsize{14}{16}\selectfont 61.5 \fontsize{10}{12}\selectfont$\pm$ 4.5 & \fontsize{14}{16}\selectfont 65.2 \fontsize{10}{12}\selectfont$\pm$ 5.4  &\fontsize{14}{16}\selectfont66.9 \\
    
    Dynamic weighting & SA.+P.G & \fontsize{14}{16}\selectfont 68.6 \fontsize{10}{12}\selectfont$\pm$ 3.5 & \fontsize{14}{16}\selectfont 67.6\fontsize{10}{12}\selectfont$\pm$ 3.5 & \fontsize{14}{16}\selectfont 78.3 \fontsize{10}{12}\selectfont$\pm$ 7.4 & \fontsize{14}{16}\selectfont 63.3 \fontsize{10}{12}\selectfont$\pm$ 5.0 & \fontsize{14}{16}\selectfont 66.6 \fontsize{10}{12}\selectfont$\pm$ 2.8  & \fontsize{14}{16}\selectfont68.9\\
    
    Dynamic weighting & SA.+G.P  & \fontsize{14}{16}\selectfont68.5 \fontsize{10}{12}\selectfont$\pm$ 2.9 & \fontsize{14}{16}\selectfont69.9\fontsize{10}{12}\selectfont$\pm$ 3.5 & \fontsize{14}{16}\selectfont73.0 \fontsize{10}{12}\selectfont$\pm$ 12.4 & \fontsize{14}{16}\selectfont63.3 \fontsize{10}{12}\selectfont$\pm$ 3.1 & \fontsize{14}{16}\selectfont67.9 \fontsize{10}{12}\selectfont$\pm$ 5.1  & \fontsize{14}{16}\selectfont68.5\\

    \cmidrule(lr){1-8}
    Dynamic weighting & IFA. & \fontsize{14}{16}\selectfont \textbf{70.5} \fontsize{10}{12}\selectfont$\pm$ 4.1 & \fontsize{14}{16}\selectfont \textbf{72.9} \fontsize{10}{12}\selectfont$\pm$ 1.9 & \fontsize{14}{16}\selectfont\textbf{80.8} \fontsize{10}{12}\selectfont$\pm$ 5.8 & \fontsize{14}{16}\selectfont \textbf{66.0} \fontsize{10}{12}\selectfont$\pm$ 5.8 & \fontsize{14}{16}\selectfont \textbf{67.5} \fontsize{10}{12}\selectfont$\pm$ 3.3 & \fontsize{14}{16}\selectfont\textbf{71.5}  \\
    \bottomrule
\end{tabular}
}
\end{table}

\textbf{Pathology feature encoder.} We evaluated various pathology feature encoders including five pretrained models: UNI~\cite{uni}, Conch~\cite{conch}, PhiKon2~\cite{Phikon2}, CTransPath~\cite{ctranspath}, and ResNet50~\cite{resnet}, as detailed in Table~\ref{table:ablation-extraction}. The UNI and CTransPath achieved the highest values of 69.6$\%$ and 76.9$\%$ on STAD and BRCA respectively. PhiKon2 performed best on BLCA (72.5$\%$) and HNSC (67.4$\%$). However, these encoders underperformed on other cohorts with suboptimal average scores. Conch demonstrated robust overall performance with leading scores in CO-READ (83.2$\%$), consistently performing well across all datasets and emerging as the most effective patch feature encoder in this study.

\subsection{Visualization}

\textbf{Cross-modality interaction}.
The pathology-genomics interactions and the impact of genomics in high- and low-risk BRCA cases are visualized in Figure~\ref{fig:visualization}. In the high-risk patient, increased expression of genes associated with cell proliferation and differentiation, such as HNF1A, CDX2, TGFBR1, ATP1B3, HNF4A, and CGA, suggests a tumor-promoting environment, while the elevated expression levels of DNA repair genes, including RECQL4, JAK1, GRK1, FCGR3A, BSX, and LECT2, indicate protective effects for the low-risk patient, enhancing the body's resistance to cancer.

\begin{table}[t]
    \centering
    \caption{\textbf{Ablation study on pathology feature encoder.} Performance of four pathology-specific pretrained models and ResNet50 is reported in C-Index ($\%$).}
    \label{table:ablation-extraction}
    \centering
    \resizebox{\linewidth}{!}{
\begin{tabular}{l|llcll|c}
    \toprule
   \fontsize{18}{20} \selectfont Patch encoder &\fontsize{18}{20} \selectfont BLCA &\fontsize{18}{20} \selectfont BRCA &\fontsize{18}{20} \selectfont CO-READ &\fontsize{18}{20} \selectfont HNSC&\fontsize{18}{20} \selectfont STAD &\fontsize{18}{20} \selectfont Mean\\
    \cmidrule(lr){1-7}
    
    \fontsize{18}{20} \selectfont UNI& \fontsize{18}{20}\selectfont 71.6\fontsize{12}{14}\selectfont$\pm$ 1.7 & \fontsize{18}{20}\selectfont 72.7\fontsize{12}{14}\selectfont$\pm$ 4.2 & \fontsize{18}{20}\selectfont 78.3\fontsize{12}{14}\selectfont$\pm$ 6.1 & \fontsize{18}{20}\selectfont 66.8\fontsize{12}{14}\selectfont$\pm$ 7.2 & \fontsize{18}{20}\selectfont \textbf{69.6}\fontsize{12}{14}\selectfont$\pm$ 6.4 & \fontsize{18}{20}\selectfont 71.8\\
    
    \fontsize{18}{20} \selectfont Conch& \fontsize{18}{20}\selectfont 67.6\fontsize{12}{14}\selectfont$\pm$ 4.6 & \fontsize{18}{20}\selectfont 75.0\fontsize{12}{14}\selectfont$\pm$ 8.2 & \fontsize{18}{20}\selectfont \textbf{83.2}\fontsize{12}{14}\selectfont$\pm$ 7.7 & \fontsize{18}{20}\selectfont 66.2\fontsize{12}{14}\selectfont$\pm$ 3.4 & \fontsize{18}{20}\selectfont 67.7\fontsize{12}{14}\selectfont$\pm$ 7.1 & \fontsize{18}{20}\selectfont \textbf{71.9}\\
    
    \fontsize{18}{20} \selectfont Phikon2& \fontsize{18}{20}\selectfont \textbf{72.5}\fontsize{12}{14}\selectfont$\pm$ 3.9  & \fontsize{18}{20}\selectfont 76.6\fontsize{12}{14}\selectfont$\pm$ 8.9 & \fontsize{18}{20}\selectfont 73.7\fontsize{12}{14}\selectfont$\pm$ 4.3 & \fontsize{18}{20}\selectfont \textbf{67.4}\fontsize{12}{14}\selectfont$\pm$ 7.8 & \fontsize{18}{20}\selectfont 68.4\fontsize{12}{14}\selectfont$\pm$ 6.5 & \fontsize{18}{20}\selectfont 71.7\\
    
    \fontsize{18}{20} \selectfont CTransPath& \fontsize{18}{20}\selectfont 66.9\fontsize{12}{14}\selectfont$\pm$ 1.9 & \fontsize{18}{20}\selectfont \textbf{76.9}\fontsize{12}{14}\selectfont$\pm$ 7.8 & \fontsize{18}{20}\selectfont 81.3\fontsize{12}{14}\selectfont$\pm$ 8.2 & \fontsize{18}{20}\selectfont 66.5\fontsize{12}{14}\selectfont$\pm$ 4.0 & \fontsize{18}{20}\selectfont 65.9\fontsize{12}{14}\selectfont$\pm$ 4.2 & \fontsize{18}{20}\selectfont 71.5\\
    \cmidrule(lr){1-7}
    \fontsize{18}{20} \selectfont ResNet50& \fontsize{18}{20}\selectfont 70.5\fontsize{12}{14}\selectfont$\pm$ 4.1 & \fontsize{18}{20}\selectfont 72.9\fontsize{12}{14}\selectfont$\pm$ 1.9 & \fontsize{18}{20}\selectfont 80.8\fontsize{12}{14}\selectfont $\pm$ 5.8 & \fontsize{18}{20}\selectfont 66.0\fontsize{12}{14}\selectfont $\pm$ 5.8 & \fontsize{18}{20}\selectfont 67.5\fontsize{12}{14}\selectfont$\pm$ 3.3 & \fontsize{18}{20}\selectfont 71.5 \\
    \bottomrule
\end{tabular}
}
    
\end{table}

\textbf{Kaplan-Meier analysis.} To validate our model's discriminative ability, we presented Kaplan-Meier curves in Figure \ref{fig:visualization}. Patients are split into high-risk and low-risk groups based on median risk scores predicted by our model, with \textit{p}-values calculated via the log-rank test. In all five datasets, \textit{p}-values below 0.05 indicate that our model effectively distinguishes between high-risk and low-risk populations.

\section{Conclusion}
In this paper, we propose a multimodal framework MRePath for cancer survival prediction. It addresses MIL-based information loss by using sheaf hypergraphs in WSIs, and pathology-genomics imbalance by employing a dynamic rebalance. Experiments on five datasets demonstrate the superior performance and effectiveness of our framework.

\textbf{Limitations.} Considering WSIs with varying numbers of patches, the same $k$-value in the hypergraph leads to different scopes, imposing a potential inconsistency in pathology slides.
In clinical applications, due to various factors, it may be challenging to obtain complete paired pathology and genomic data. In some cases, low-quality data or even missing modalities can significantly impact our modality rebalancing process, highlighting the need for greater robustness of our framework.
We only considered a single WSI or a set of genomic information. However, there may be multiple WSIs and more extensive genomic data, which introduces the need for more sophisticated rebalancing and fusion strategies.

\clearpage
\newpage

\bibliographystyle{named}
\bibliography{WSI}

\clearpage
\newpage
In this supplementary material, we provide a detailed overview of the framework implementation and experimental specifics. This document includes:

\begin{itemize}
\item \textbf{Model Implementation Details:} A comprehensive explanation of the entire process and module implementations, outlining the architecture and algorithms used, as presented in Section \ref{section:id}.
\item \textbf{Dataset Information:} Detailed descriptions of the datasets employed in our experiments, including their sources and distribution, as provided in Section \ref{setion:dataset}.
\item \textbf{Experimental Data and Visualization of Results:} Detailed experimental results and visualizations are included to clearly illustrate the outcomes and support our research, as outlined in Section \ref{section:er}.
\end{itemize}

\section{Implementation Details}
\label{section:id}
We introduce three crucial algorithms about the implementation of our model. 

\textbf{MRePath framework flow:} This algorithm details each stage of the process, including feature extraction, hypergraph learning, and modality rebalance, as shown in Algorithm~\ref{alg:MRePath}. It provides insight into how pathology and genomic data are integrated and processed to enable precise survival predictions.

\textbf{Hypergraph construction:} Tailored specifically for pathology data, this algorithm elucidates the method used to capture spatial and structural interrelations within whole slide images (WSIs), as illustrated in Algorithm~\ref{alg:hypergraph}. It enhances the representation of contextual and hierarchical nuances that are critical for sophisticated analysis. 

\textbf{Dynamic weighting mechanism:} This involves calculating mono-confidence and holo-confidence measures to adjust modality weights, thereby rebalancing the pathology-genomics imbalance, as detailed in Algorithm~\ref{alg:dynamic_weighting}.

\begin{figure}[t]
 \centering
 \includegraphics[width=1\linewidth]{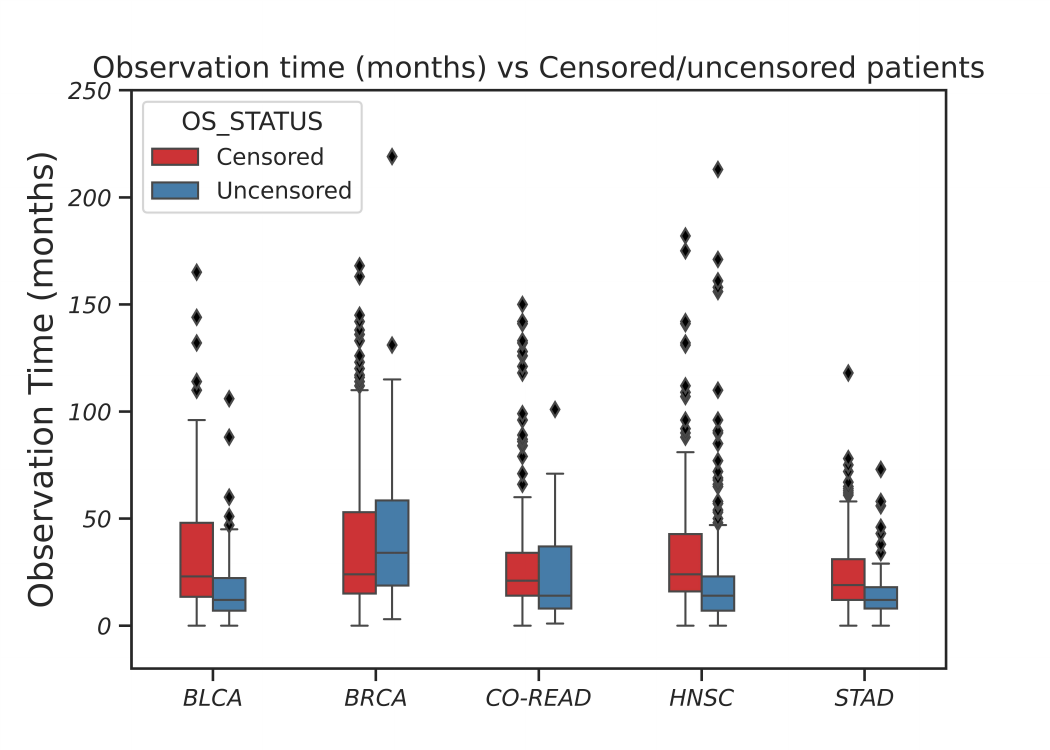}
 \caption{Visualization for the distribution of each dataset using box plots based on censorship status. Black diamonds represent outliers.}
 \label{fig:dataset}
\end{figure}

\begin{algorithm*}[h]
\caption{MRePath Framework Flow}
\label{alg:MRePath}

\KwIn{WSIs, genomics profiles}
\SetKwBlock{StageOne}{Stage 1: Feature Extraction}{end}
\SetKwBlock{StageTwo}{Stage 2: Hypergraph Construction}{end}
\SetKwBlock{StageThree}{Stage 3: Modality Rebalance}{end}

\StageOne{
    \textbf{Input:} WSIs, genomics profiles\;
    \quad Extract features \(\mathbf{P}\) using pathology encoder from WSI\; 
    \quad Extract features \(\mathbf{G}\) using genomics encoder from genomics\; 
    \textbf{Output:} Pathology features \(\mathbf{P}\), Genomics features \(\mathbf{G}\)
}

\StageTwo{
    \textbf{Input:} Pathology features \(\mathbf{P}\)\;
    \quad Construct hypergraph $\mathcal{G}=\{\mathcal{E},\mathcal{V}\}$\;
    \quad Apply sheaf hypergraph to enhance feature representations with contextual and hierarchical details\;
    \textbf{Output:} High-order features \(\mathbf{P}_h\)\;
}

\StageThree{
    \textbf{Input:} High-order features \(\mathbf{P}_h\), Raw pathology features \(\mathbf{P}\), Genomics features \(\mathbf{G}\)\;
    \quad Dynamic weighting of \(\mathbf{P}\), \(\mathbf{G}\) to obtain weights \(w_p\) and \(w_g\) \;
    \quad Rebalance the features using $w_p$, $w_g$ to obtain balanced features 
  	\(\mathbf{P}_w = w_p \cdot 
  	\mathbf{
  		P}_h
  	\) and 
  	\(\mathbf{
  		G}_w = w_g 
  	\cdot 
  	\mathbf{
  		G}\) \;
    \quad Use interactive alignment fusion to integrate features, resulting in integrated features 
 	\(\mathbf{
 		P}_f
 	\) and 
 	\(\mathbf{
 		G}_f
 	\) \;
    \quad Predict final risk outcomes based on 
 	\(\mathbf{
 		P}_f
 	\) and 
 	\(\mathbf{
 		G}_f
 	\)\;
    \textbf{Output:} Survival risk
}
\Return Survival risk
\end{algorithm*}

\section{Datasets}
\label{setion:dataset}
The datasets used in our experiments are as follows: Bladder Urothelial Carcinoma (BLCA) with 384 samples, Breast Invasive Carcinoma (BRCA) with 968 samples, Colon and Rectum Adenocarcinoma (CO-READ) with 298 samples, Head and Neck Squamous Cell Carcinoma (HNSC) with 392 samples, and Stomach Adenocarcinoma (STAD) with 317 samples. These datasets were sourced from The Cancer Genome Atlas (TCGA) repository\footnote{\url{https://portal.gdc.cancer.gov/}}, which provides comprehensive genomic profiles for various cancer types.

To illustrate the distribution of survival times within these datasets, we utilized box plots, as shown in Fig.\ref{fig:dataset}. The division of the datasets was guided by the approach outlined in MCat \cite{Mcat}.

\section{Experimental results}
\label{section:er}
\subsection{Ablation study on hypergraph learning.}
The detailed experimental data of the ablation study on hypergraph learning is summarized in Table~\ref{table:hl}. We first focused on the impact of different graph structures on the module's performance, including the use of MLP for feature aggregation (without a graph structure), Graph Attention Network (GAT), and Graph Convolutional Network (GCN). Additionally, we examined the effect of different hyperedge types on the module's performance, specifically analyzing topological-based hyperedges \(\mathcal{E}_T\) and feature-based hyperedges \(\mathcal{E}_F\).

\begin{table*}[h]
 \centering

 \resizebox{\linewidth}{!}{
 \begin{tabular}{c|ccc|ccc|ccc|ccc|ccc}
 \cmidrule(lr){1-16}
 GNN& \multicolumn{3}{c|}{BLCA} & \multicolumn{3}{c|}{BRCA} & \multicolumn{3}{c|}{CO-READ} & \multicolumn{3}{c|}{HNSC} & \multicolumn{3}{c}{STAD} \\ 
 \cmidrule(lr){1-16}
 Model & / & GAT & GCN 
  & / & GAT & GCN 
  & / & GAT & GCN 
  & / & GAT & GCN 
  & / & GAT & GCN \\ 
 \cmidrule(lr){1-16}
 Split 0 
  & 63.3 & 73.9 & 65.2 
  & 71.8 & 65.9 & 74.1 
  & 66.7 & 74.0 & 67.1 
  & 65.8 & 61.0 & 58.4 
  & 67.9 & 64.8 & 74.3 \\ 
 Split 1 
  & 63.3 & 69.4 & 65.9 
  & 66.0 & 67.0 & 65.3 
  & 56.9 & 60.9 & 73.5 
  & 59.7 & 61.2 & 61.1 
  & 58.1 & 66.0 & 62.4 \\ 
 Split 2 
  & 59.3 & 68.7 & 72.3 
  & 77.4 & 70.6 & 70.1
  & 77.9 & 75.0 & 87.5 
  & 68.7 & 71.5 & 58.4 
  & 63.7 & 73.0 & 63.3 \\ 
 Split 3 
  & 59.2 & 63.3 & 74.2 
  & 73.1 & 67.9 & 70.4 
  & 78.9 & 74.5 & 74.0 
  & 59.9 & 56.7 & 63.3
  & 63.5 & 60.4 & 73.8\\ 
 
  Split 4 
  
  & 67.6 & 71.8 & 69.6 
  & 69.4 & 67.2 & 67.7
  & 69.1 & 70.7 & 61.2 
  & 65.5 & 65.5 & 59.5 
  & 55.0 & 58.4 & 60.8
  \\ 
 \cmidrule(lr){1-16}
 Mean 
  & 62.5 & 69.5 & 69.0 
  & 71.5 & 67.7 & 69.5 
  & 69.9 & 71.0 & 72.6
  & 63.9 & 62.6& 60.1
  & 61.6 & 64.5 &66.9 \\ 
 \fontsize{8}{10}\selectfont std & \fontsize{8}{10}\selectfont3.5 &\fontsize{8}{10}\selectfont 6.3 & \fontsize{8}{10}\selectfont3.6
  &\fontsize{8}{10}\selectfont4.2&\fontsize{8}{10}\selectfont1.8&\fontsize{8}{10}\selectfont3.3
  &\fontsize{8}{10}\selectfont9.0&\fontsize{8}{10}\selectfont5.9&\fontsize{8}{10}\selectfont9.8
  &\fontsize{8}{10}\selectfont3.9&\fontsize{8}{10}\selectfont5.4&\fontsize{8}{10}\selectfont2.1
  &\fontsize{8}{10}\selectfont5.0 &\fontsize{8}{10}\selectfont5.7&\fontsize{8}{10}\selectfont6.6\\

\cmidrule(lr){1-16}
Model & HGNN & Sheaf ($\mathcal{E}_T$) & Sheaf ($\mathcal{E}_F$)
& HGNN & Sheaf ($\mathcal{E}_T$) & Sheaf ($\mathcal{E}_F$)
& HGNN & Sheaf ($\mathcal{E}_T$) & Sheaf ($\mathcal{E}_F$)
& HGNN & Sheaf ($\mathcal{E}_T$) & Sheaf ($\mathcal{E}_F$)
& HGNN & Sheaf ($\mathcal{E}_T$) & Sheaf ($\mathcal{E}_F$) \\ \cmidrule(lr){1-16}
Split 0 & 66.5 & 63.4 & 65.7
& 78.6 & 66.3 & 67.6
& 70.7 & 74.0 & 70.7
& 63.8 & 61.3 & 72.9
& 73.2 & 72.9 & 68.2 \\
Split 1 & 67.8 & 69.8 & 70.0
& 74.2 & 70.1 & 75.8
& 77.3 & 68.8 & 75.3
& 66.6 & 60.9 & 58.8
& 67.0 & 65.2 & 64.8 \\
Split 2 & 70.2 & 70.4 & 69.1
& 75.4 & 72.8 & 70.6
& 87.5 & 90.4 &89.4&68.6&64.3&73.0&61.8&64.7&64.3\\
Split 3 & 79.4 & 70.5 & 70.0
& 64.2 & 72.4 & 72.3
& 82.1 & 79.7 &78.3&63.1&61.5&59.9&68.1&65.4&68.9\\
Split 4 & 72.1 & 67.9 & 71.8
& 69.4 & 69.1 & 66.6
& 82.4 & 72.7 &83.0&62.6&64.3&60.9&60.2&57.7&60.5\\
\cmidrule(lr){1-16}
Mean &69.2&68.4&69.3
 &72.4&70.1&70.1
 &80.0&77.1&79.4
 &64.9&62.4&65.1
 &66.1&65.3&65.3\\
\fontsize{8}{10}\selectfont
std 
 &\fontsize{8}{10}\selectfont2.2&\fontsize{8}{10}\selectfont2.9&\fontsize{8}{10}\selectfont2.2
 &\fontsize{8}{10}\selectfont5.6&\fontsize{8}{10}\selectfont2.6&\fontsize{8}{10}\selectfont3.7
 &\fontsize{8}{10}\selectfont6.3&\fontsize{8}{10}\selectfont8.4&\fontsize{8}{10}\selectfont7.2
 &\fontsize{8}{10}\selectfont2.5&\fontsize{8}{10}\selectfont1.7&\fontsize{8}{10}\selectfont7.2
 &\fontsize{8}{10}\selectfont5.2&\fontsize{8}{10}\selectfont5.3&\fontsize{8}{10}\selectfont3.4\\
\bottomrule

\end{tabular}
}

 \caption{Ablation study results on hypergraph learning module performance of different graph structures, including MLP (without a graph), GAT, and GCN., and various hyperedge types, such as topological-based hyperedges $\mathcal{E}_T$ and feature-based hyperedges $\mathcal{E}_F$, is reported in C-index ($\%$).}
 \label{table:hl}
\end{table*}

\subsection{Ablation studies on hyperedge construction threshold \textit{k}}
The detailed experimental data of ablation study on hyperedge construction is summarized in Table~\ref{table:k_results}.
We compared the experimental results of different hyperedge thresholds across five datasets and reported for five data splits.

For the BLCA dataset, the mean C-index scores suggest that a threshold of \(k\) = 9 yields the best performance with a mean score of 70.5\%, while other thresholds result in lower averages. The standard deviation for \(k\) = 9 is relatively low at 4.0\%, indicating inconsistent performance across splits. In contrast, higher variability is observed for other thresholds, particularly \(k\) = 4.
In the BRCA dataset, both \(k\) = 9 and \(k\) = 48 achieve a high mean scores of 72.9\% and 71.5\%, respectively. However, \(k\) = 48 shows higher variability with a standard deviation of 5.6\% compared to \(k\) = 9, suggesting that while both thresholds perform well on average, \(k\) = 9 offers more consistent results.
The CO-READ dataset exhibits the highest mean performance with a threshold of \(k\) = 48, achieving an average C-index of 81.1\%. However, this configuration also has a high standard deviation of 10.0\%, indicating significant variability across splits. The threshold \(k\) = 9 provides a slightly lower mean score but with less variability.
For HNSC, the results are more balanced across different thresholds, with no single value significantly outperforming others in terms of mean C-index scores. The variability is moderate across all thresholds, suggesting consistent but modest improvements in performance.
In the STAD dataset, the threshold \(k\) = 4 achieves the highest mean score of 69.0\%. For $k$ = 9 and $k$ = 48, both achieve the same mean score; however, the standard deviation for $k$ = 9 is significantly smaller than that for ($k$ = 48) (3.3\% versus 9.7\%). This indicates that different $k$ values impact not only the overall performance but also the consistency across different splits.

Overall, our resluts highlights how varying the hyperedge construction threshold impacts model performance differently across datasets. 
Proper tuning of $k$ enhances model performance by capturing meaningful relationships while avoiding information homogenization ($k$ = 49) or oversimplification ($k$ = 0).

\begin{table*}[t]
 \centering
 \resizebox{\linewidth}{!}{
 \begin{tabular}{c|ccccc|ccccc|ccccc|ccccc|ccccc}
 \toprule
 & \multicolumn{5}{c|}{BLCA} & \multicolumn{5}{c|}{BRCA} & \multicolumn{5}{c|}{CO-READ} & \multicolumn{5}{c|}{HNSC} & \multicolumn{5}{c}{STAD} \\ 
 \cmidrule(lr){2-6} \cmidrule(lr){7-11} \cmidrule(lr){12-16} \cmidrule(lr){17-21} \cmidrule(lr){22-26}
 $k$ & 0 & 4 & 9 & 24 & 48 
  & 0 & 4 & 9 & 24 & 48
  & 0 & 4 & 9 & 24 & 48
  & 0 & 4 & 9 & 24 & 48
  & 0 & 4 & 9 & 24 & 48\\ 
 \midrule
 Split 0 & 63.3 & 68.0 & 64.6 & 67.8 & 62.1 
   & 71.8 & 70.3 & 70.4 & 67.3 & 78.8
   & 66.7 & 84.3 & 85.0 & 72.0 & 93.5
   & 65.8 & 66.1 & 63.9 & 66.4 & 69.6 
   & 67.9 & 77.8 & 67.4 & 75.7 & 81.0 \\

 Split 1 & 63.3 & 63.7 & 69.2 & 67.2 & 65.6
   & 66.0 & 64.4 & 73.8 & 71.2 & 68.2
   & 56.9 & 62.3 & 75.0 & 75.3 & 75.5
   & 59.7 & 63.3 & 73.4 & 61.4 & 56.4
   & 58.1 & 69.0 & 69.4 & 63.6 & 69.8
   \\
 Split 2 & 59.3 & 69.2 & 72.5 & 61.7 & 74.1
   & 77.4 & 65.7 & 75.4 & 71.9 & 75.7
   & 71.9 & 72.1 & 87.5 & 82.7 & 90.4
   & 68.7 & 63.3 & 70.9 & 67.1 & 69.1
   & 63.7 & 65.0 & 68.3 & 64.2 & 63.6
   \\

 Split 3 & 59.2 & 60.5 & 75.5 & 63.4 & 69.7
   & 73.1 & 70.0 & 72.1 & 72.0 & 75.7
   & 78.9 & 75.9 & 81.8 & 69.1 & 72.9
   & 59.9 & 62.7 & 61.5 & 60.4 & 65.3
   & 63.5 & 71.5 & 70.4 & 65.4 & 68.9
   \\
 Split 4 & 67.6 & 60.8 & 70.8 & 60.8 & 61.7
   & 69.4 & 69.1 & 73.0 & 70.0 & 72.1
   & 69.1 & 66.1 & 74.6 & 73.9 & 73.3
   & 65.5 & 61.4 & 60.2 & 66.1 & 68.9
   & 55.0 & 61.7 & 61.8 & 57.7 & 54.4
   \\
 \cmidrule(lr){1-26}
 Mean & 62.5 & 64.4 & 70.5 & 64.0 & 64.2
  & 71.5 & 65.9 & 72.9 & 70.5 & 71.5
  & 69.9 & 72.1 & 80.8 & 74.9 & 81.1
  & 63.9 & 63.4 & 66.0 & 64.3 & 65.8
  & 61.6 & 69.0 & 67.5 & 65.3 & 67.5
 
 \\
 \fontsize{8}{10}\selectfont std &\fontsize{8}{10}\selectfont 3.5 & \fontsize{8}{10}\selectfont4.0 & \fontsize{8}{10}\selectfont4.0 & \fontsize{8}{10}\selectfont3.3 & \fontsize{8}{10}\selectfont2.2
  & \fontsize{8}{10}\selectfont4.2 & \fontsize{8}{10}\selectfont4.6 & \fontsize{8}{10}\selectfont1.9 &\fontsize{8}{10}\selectfont 2.0 &\fontsize{8}{10}\selectfont 5.6
  & \fontsize{8}{10}\selectfont9.0 & \fontsize{8}{10}\selectfont8.6 & \fontsize{8}{10}\selectfont5.8 &\fontsize{8}{10}\selectfont 5.1 & \fontsize{8}{10}\selectfont10.0
  & \fontsize{8}{10}\selectfont3.9 & \fontsize{8}{10}\selectfont1.7 & \fontsize{8}{10}\selectfont5.8 & \fontsize{8}{10}\selectfont3.1 & \fontsize{8}{10}\selectfont5.5
  & \fontsize{8}{10}\selectfont6.0 \fontsize{8}{10}\selectfont& 6.2 & \fontsize{8}{10}\selectfont3.3 & \fontsize{8}{10}\selectfont6.5 & \fontsize{8}{10}\selectfont9.7
   \\ 
 \bottomrule
 \end{tabular}
 }
 \caption{Ablation study results on hyperedge construction threshold $k$. Performance of various similarity thresholds ($k$ =0, 5, 9, 25, 49)
is reported in C-Index ($\%$).}
 \label{table:k_results}
\end{table*}

\begin{algorithm}[h]
\caption{Hypergraph Construction for Pathology}
\label{alg:hypergraph}

\KwIn{Pathology $\mathbf{P}$, Hyperedge construction threshold $k$, pathology features $f_v$ and patch coordinates $c_{p}$}

Initialize set of vertices $\mathcal{V} = \{v_1, v_2, \dots, v_N\}$\;

\ForEach{patch $v_i \in \mathcal{V}$}{
    Determine topological neighbors (Eq. 2)\;
    Form topological hyperedge $\mathcal{E}_T$\;
}

\ForEach{patch $v_i \in \mathcal{V}$}{
    Determine feature-based neighbors (Eq. 3)\;
    Form feature-based hyperedge $\mathcal{E}_F$\;
}

Construct final hyperedge set: $\mathcal{E} = \mathcal{E}_T \cup \mathcal{E}_F$\;

\Return $\mathcal{G} = \{\mathcal{V}, \mathcal{E}\}$\;

\end{algorithm}

\begin{algorithm}[t]
\caption{Dynamic Weighting for Modality Fusion}
\label{alg:dynamic_weighting}

\KwIn{Pathology features \(\mathbf{P}\), Genomics features \(\mathbf{G}\)}

\SetKwBlock{StageOne}{Stage 1: Mono-confidence Calculation}{end}
\SetKwBlock{StageTwo}{Stage 2: Holo-confidence Calculation}{end}
\SetKwBlock{StageThree}{Stage 3: Final Weight Calculation}{end}

\StageOne{
    Compute mono-confidence for pathology: \(w_p^{m} = \mathbf{P} \Phi_p\)\;
    Compute mono-confidence for genomics: \(w_g^{m} = \mathbf{G} \Phi_g\)\;
}

\StageTwo{
    Compute holo-confidence for pathology: \(w_p^{h} = \frac{\log(w_p^{m})}{\log(w_p^{m} \cdot w_g^{m})}\)\;
    Compute holo-confidence for genomics: \(w_g^{h} = \frac{\log(w_g^{m})}{\log(w_p^{m} \cdot w_g^{m})}\)\;
}

\StageThree{
    Combine mono- and holo-confidence scores\;
    Apply softmax to obtain final weights: \(w_p, w_g = \phi(w_p^{m} + w_p^{h}, w_g^{m} + w_g^{h})\)\;
}

\Return Final weights \(w_p, w_g\)\;

\end{algorithm}

\subsection{Ablation study on modality rebalance.}
The detailed experimental data of ablation study on modality rebalance is detailed in Table~\ref{table:weight}.
We first investigated the experimental results of using fixed weights for rebalance, which included equal weights ($w_p=0.5,w_g=0.5$), pathology-biased ($w_p=0.7,w_g=0.3$), and genomics-biased ($w_p=0.3,w_g=0.7$) configurations.
To determine the impact of different modalities on the final outcome, we adjusted the contribution values to 0.05 ($w_p=0.05, w_g=0.95$ and $w_p=0.95, w_g=0.05$), as shown in Table~\ref{table:weight}. 
We then examined the role of different layers in the modality mixing process.

\begin{table*}[t]
\centering
\resizebox{\linewidth}{!}{
\begin{tabular}{c|ccc|ccc|ccc|ccc|ccc}
\toprule
Dataset &\multicolumn{3}{c|}{BLCA} & \multicolumn{3}{c|}{BRCA} & \multicolumn{3}{c|}{CO-READ} & \multicolumn{3}{c|}{HNSC} & \multicolumn{3}{c}{STAD} \\ 
\cmidrule(lr){1-16}
$w_p$, $w_g$& 0.5, 0.5 & 0.7, 0.3 & 0.3, 0.7
& 0.5, 0.5 & 0.7, 0.3 & 0.3, 0.7
& 0.5, 0.5 & 0.7, 0.3 & 0.3, 0.7
& 0.5, 0.5 & 0.7, 0.3 & 0.3, 0.7 
& 0.5, 0.5 & 0.7, 0.3 & 0.3, 0.7
\\ \cmidrule(lr){1-16}
Split 0 &63.9&64.1&64.3& 65.7 & 66.0 & 68.0
& 65.0 & 80.5 & 66.7
& 66.6 & 66.6 & 66.9
& 63.3 & 70.5 & 72.9 \\
Split 1 &72.3&68.4&65.6& 70.2 & 69.8 & 70.3
& 61.3 & 68.3 & 75.9
& 67.9 & 71.9 & 60.2
& 66.4 & 70.5 & 64.2 \\
Split 2 &72.1&69.5&65.6& 64.9 & 68.0 & 68.3
& 82.7 & 92.3 & 82.7
& 66.1 & 61.7 &63.1&71.7&65.4&69.7
\\
Split 3&69.8&70.8&72.1&71.0&77.6&79.1
&74.8&77.0&76.2&57.9&58.6&60.7&71.5&75.6&71.3
\\
Split 4&68.5&68.7&69.8&64.3&63.1&69.3
&70.3&63.0&77.0&66.9&69.5&69.6&59.5&54.4&62.2
\\
\cmidrule(lr){1-16}
Mean&69.3&68.3&68.6&67.2&68.9&71.0&70.8&76.2&75.7&65.1&65.6&64.1&66.5&67.3&68.1\\
\fontsize{8}{10}\selectfont std &\fontsize{8}{10}\selectfont 3.0&\fontsize{8}{10}\selectfont2.5&\fontsize{8}{10}\selectfont3.5&\fontsize{8}{10}\selectfont3.1&\fontsize{8}{10}\selectfont5.5&\fontsize{8}{10}\selectfont4.6&\fontsize{8}{10}\selectfont8.4&\fontsize{8}{10}\selectfont11.3&\fontsize{8}{10}\selectfont5.7&\fontsize{8}{10}\selectfont4.1&\fontsize{8}{10}\selectfont5.5&\fontsize{8}{10}\selectfont4.1&\fontsize{8}{10}\selectfont5.3&\fontsize{8}{10}\selectfont8.0&\fontsize{8}{10}\selectfont4.7\\

 \cmidrule(lr){1-16}
 Layers & P.G+G.P & SA.+P.G & SA.+G.P 
  & P.G+G.P & SA.+P.G & SA.+G.P 
  & P.G+G.P & SA.+P.G & SA.+G.P 
  & P.G+G.P & SA.+P.G & SA.+G.P 
  & P.G+G.P & SA.+P.G & SA.+G.P
  \\
 \cmidrule(lr){1-16}
 Split 0 & 62.7 &63.3 &65.1
  & 69.0 & 64.6 & 68.6 
  & 69.9 & 72.4 & 72.0 
  & 61.3 & 64.0 & 66.3 
  & 64.9 & 71.0 & 67.7 \\ 
 Split 1 & 65.3 & 67.2 & 68.3
  & 69.6 & 69.3 & 74.3 
  & 66.1 & 68.9 & 62.3 
  & 59.2 & 57.3 & 65.2
  & 60.1 & 64.0 & 66.6\\
 Split 2 & 69.4 & 72.0 & 68.9
  & 65.5 & 63.3 & 66.9
  & 85.6 & 86.5 & 94.2
  & 58.5 & 70.6 & 62.7
  & 70.4 & 66.9 & 73.14\\
 Split 3 & 72.0 & 70.8 & 73.0
  & 71.8 & 71.8 & 73.1
  & 77.0 & 82.7 & 69.4
  & 59.4 & 60.2 & 58.8
  & 70.8 & 66.9 & 71.9\\
 Split 4 & 68.7 & 69.6 & 67.5
  & 69.6 & 68.8 & 66.7
  & 57.0 & 81.2 & 67.3
  & 69.3 & 60.2 & 58.8
  & 59.6 & 64.4 & 60.2\\
 \cmidrule(lr){1-16}
 Mean & 67.6&68.6
 &68.6&69.1&67.6&69.9&71.1&78.3&73.0
 &61.5&63.3&63.3
 &70.7&66.0&71.3\\ 
 \fontsize{8}{10}\selectfont std&\fontsize{8}{10}\selectfont3.6&\fontsize{8}{10}\selectfont3.5&\fontsize{8}{10}\selectfont2.9&\fontsize{8}{10}\selectfont2.3&\fontsize{8}{10}\selectfont3.5&\fontsize{8}{10}\selectfont3.5&\fontsize{8}{10}\selectfont10.8&\fontsize{8}{10}\selectfont7.4&\fontsize{8}{10}\selectfont12.4&\fontsize{8}{10}\selectfont4.5&\fontsize{8}{10}\selectfont5.0&\fontsize{8}{10}\selectfont3.1&\fontsize{8}{10}
 \selectfont5.4&\fontsize{8}{10}
 \selectfont2.8&\fontsize{8}{10}
 \selectfont5.1\\
 \bottomrule
 \end{tabular}
 }
 \caption{Ablation study on modality rebalance. Performance of various modality weights and different fusion strategies is reported in C-index ($\%$). ``IFA" denotes the proposed Interactive Alignment Fusion. ``P.G", ``G.P", and ``SA." denote pathology-guided and gene-guided co-attention layers, and self-attention for genomics respectively.}
 \label{mr:2}
\end{table*}
\begin{table*}[t]
 \centering
 \resizebox{\linewidth}{!}{
 \begin{tabular}{c|ccccc|ccccc|ccccc|ccccc|ccccc}
 
 \toprule
 & \multicolumn{5}{c|}{BLCA} & \multicolumn{5}{c|}{BRCA} & \multicolumn{5}{c|}{CO-READ} & \multicolumn{5}{c|}{HNSC} & \multicolumn{5}{c}{STAD} \\ 
 \cmidrule(lr){1-26}
 Model & U. & C. & P. & T. & R. 
  & U. & C. & P. & T. & R.
  & U. & C. & P. & T. & R.
  & U. & C. & P. & T. & R.
  & U. & C. & P. & T. & R.\\ 
 \cmidrule(lr){1-26}
 Split 0 & 71.8 & 72.1 & 69.6 & 68.8 & 64.6 
   & 78.3	& 79.3	& 88.8	& 87.5	& 70.4 
   & 83.7	& 82.1	& 80.1	& 84.5	& 85.0 
   & 74.9	& 65.6	& 64.3	& 59.7	& 63.9 
   & 72.9	& 78.5	& 71.9	& 72.5	& 71.9 \\

 Split 1 & 72.0 & 64.0 & 69.3 & 65.6 & 69.2 
   & 67.8	& 69.9	& 82.1	& 70.6	& 73.8 
   & 68.9	& 76.7	& 69.1	& 71.3	& 75.0 
   & 64.8	& 67.8	& 65.8	& 67.3	& 73.4 
   & 68.8	& 66.0	& 67.4	& 63.8	& 69.4
   \\
 Split 2 & 74.1 & 64.9 & 78.1 & 68.1 & 72.5 
   & 70.4	& 87.4	& 62.7	& 81.9	& 75.4 
   & 78.9	& 96.2	& 75.0	& 92.3	& 87.5 
   & 72.4	& 70.0	& 74.8	& 70.6	& 70.9 
   & 66.3	& 62.1	& 70.4	& 67.4	& 68.3 
   \\
 Split 3 & 70.6 & 64.0 & 75.0 & 64.7 & 75.5 
   & 75.4	& 67.5	& 78.3	& 75.3	& 72.1 
   & 76.4	& 82.9	& 70.5	& 79.7	& 81.8 
   & 56.4	& 60.8	& 56.7	& 67.3	& 61.5 
   & 78.5	& 70.8	& 74.6	& 63.9	& 70.4 
   \\
 Split 4 & 69.5 & 73.1 & 70.4 & 67.7 & 70.8 
   & 71.5	& 70.9	& 74.2	& 69.1	& 73.0 
   & 83.6	& 78.2	& 73.9	& 77.6	& 74.6 
   & 65.9	& 66.6	& 75.3	& 67.8	& 60.2 
   & 61.8	& 61.3	& 57.8	& 62.0	& 61.8 
   \\
 \cmidrule(lr){1-26} 
 Mean & 71.6 & 67.6 & 72.5 & 66.9 & 70.5 
   & 72.7	& 75.0	& 76.6	& 76.9	& 72.9 
   & 78.3	& 83.2	& 73.7	& 81.3	& 80.8 
   & 66.9	& 66.2	& 67.4	& 66.5	& 66.0 
   & 69.6	& 67.7	& 68.4	& 65.9	& 67.5
   \\
 
 \fontsize{8}{10}\selectfont std &\fontsize{8}{10}\selectfont 1.7 &\fontsize{8}{10}\selectfont 4.6 &\fontsize{8}{10}\selectfont 3.9 &\fontsize{8}{10}\selectfont 1.9 &\fontsize{8}{10}\selectfont 4.1 
   &\fontsize{8}{10}\selectfont 4.2	&\fontsize{8}{10}\selectfont 8.2	&\fontsize{8}{10}\selectfont 8.9	&\fontsize{8}{10}\selectfont 7.8	&\fontsize{8}{10}\selectfont 1.9 
   &\fontsize{8}{10}\selectfont 6.1	&\fontsize{8}{10}\selectfont 7.7	&\fontsize{8}{10}\selectfont 4.3	&\fontsize{8}{10}\selectfont 8.2	&\fontsize{8}{10}\selectfont 5.8 
   &\fontsize{8}{10}\selectfont 7.2	&\fontsize{8}{10}\selectfont 3.4	&\fontsize{8}{10}\selectfont 7.8	&\fontsize{8}{10}\selectfont 4.0	&\fontsize{8}{10}\selectfont 5.8 
   &\fontsize{8}{10}\selectfont 6.4	&\fontsize{8}{10}\selectfont 7.1	&\fontsize{8}{10}\selectfont 6.5	&\fontsize{8}{10}\selectfont 4.2	&\fontsize{8}{10}\selectfont 3.3
   
   \\ 
 \bottomrule
 \end{tabular}
 }
 \caption{Ablation study on pathology feature encoder. Performance of four pathology-specific pretrained models and ResNet50 is reported in C-index ($\%$). ``U.", ``C.", ``P.", ``T.", and ``R." is UNI, Conch, Phikon2, CTransPath, and ResNet50, respectively.}
 \label{fe}
\end{table*}

\subsection{Ablation study on pathology feature encoder.}
The detailed experimental data of ablation study on pathology feature encoder is detailed in Table~\ref{fe}.
For the BLCA dataset, the mean C-index scores suggest that the models perform relatively similarly, with Phikon2 achieving a highest average score of 72.5\%. However, the standard deviation indicates some variability in performance across splits, particularly for Conch and Phikon2. 
In the BRCA dataset, CTranspath performed best with a score of 76.9\%, Phikon2 stands out with a notably high mean score of 76.6$\%$, although it also exhibits significant variability as shown by its high standard deviation of 8.9\%. This suggests that while Phikon2 can perform well on this dataset, its results may not be consistent across different splits.
The CO-READ dataset shows generally high performance across all models, with Conch achieving the highest mean score of 83.2$\%$ and a relatively moderate standard deviation of 7.7$\%$, which indicates both strong and fairly consistent performance. 
For HNSC, all models show lower mean scores compared to other datasets, with Phikon2 achieving the highest average at 67.4$\%$.
In the STAD dataset, there is more variability in performance among the models. UNI achieves the highest mean score at 69.6$\%$, but with a considerable standard deviation of 6.4$\%$, indicating variability in outcomes across different splits.

\begin{figure}[t]
    \centering
    \includegraphics[width=1.0\linewidth]{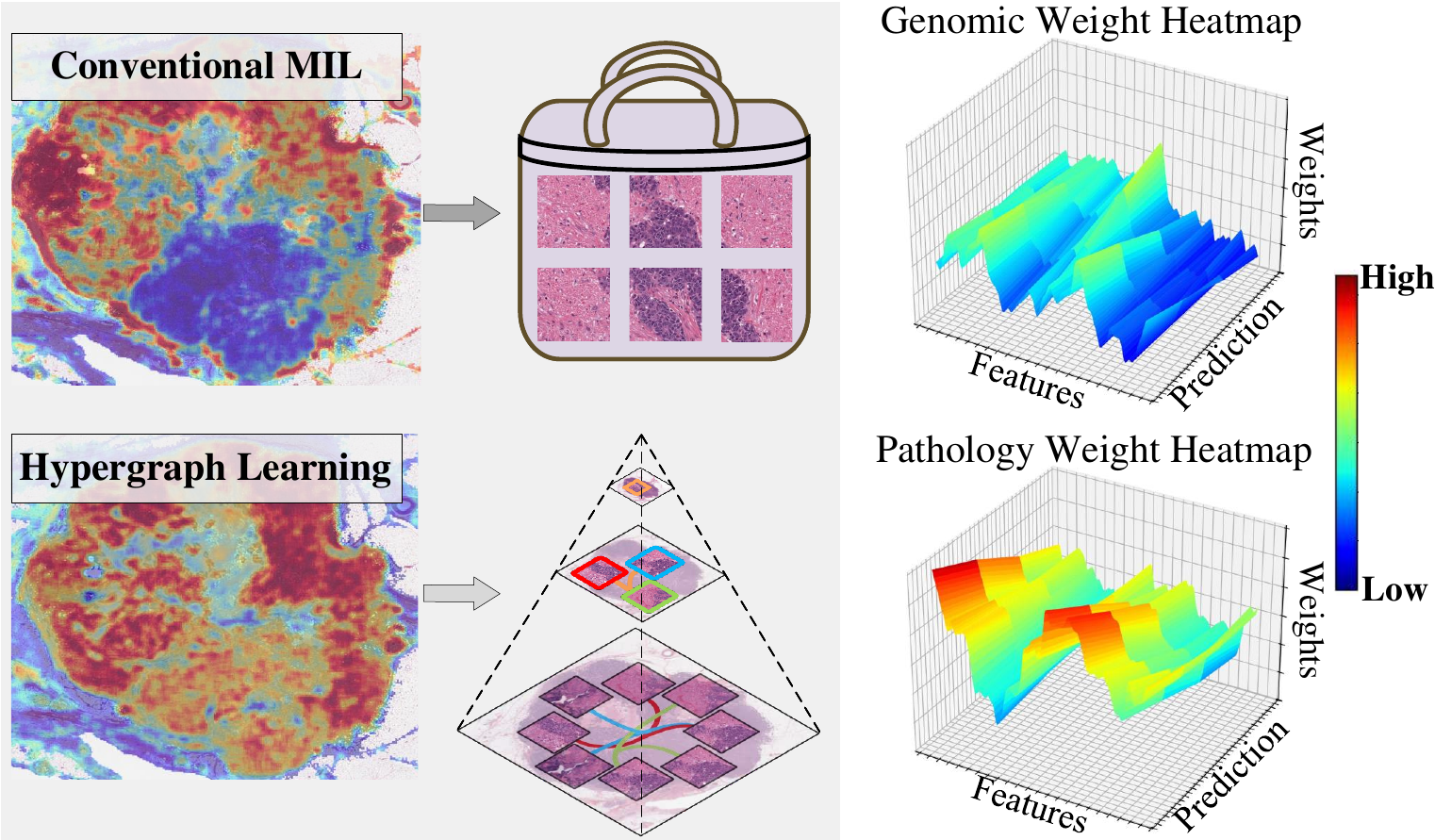}
    \caption{\textbf{a)}, Density of modality weights, where purple indicates WSI (pathology) and green indicates genomics. \textbf{b)}, Visualization of the prediction layer weights after applying modality rebalance.}
    \label{fig:heatmap}
\end{figure}

\begin{figure*}[h]
 \centering
 \includegraphics[width=1\linewidth]{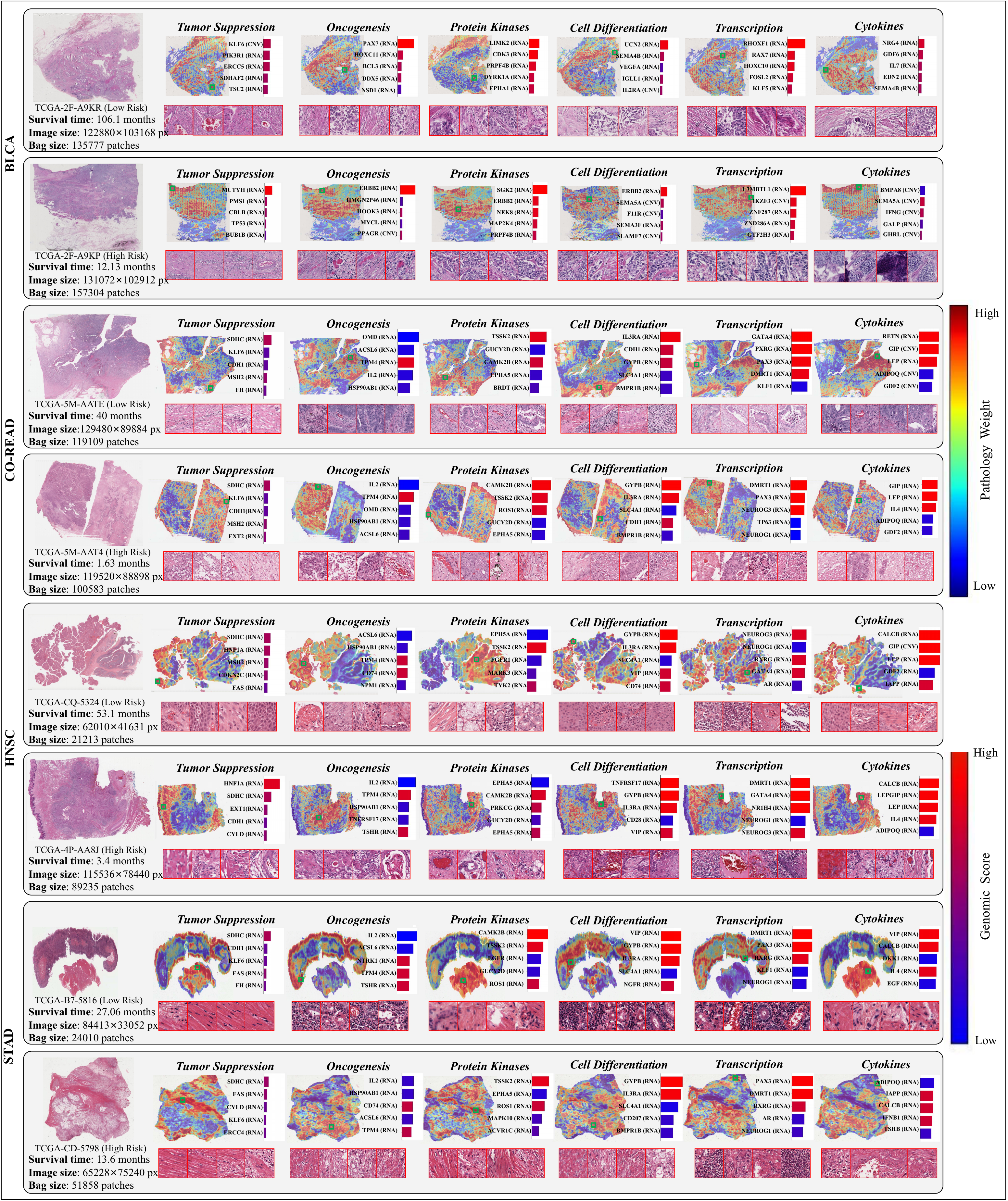}
 \caption{For visualization of the BLCA, CO-READ, HNSC and STAD dataset, one low-risk patient (Top) and one high-risk patient (Bottom) were selected from the these dataset, and the interaction between pathology and genomics and the impact of genomics were visualized}
 \label{fig:blca}
\end{figure*}

\subsection{Ablation study on genomics aggregation}
We treat the six genomes as a fully connected graph and have experimented with using different graph neural networks to aggregate genetic information.
When using GCN, GAT for modeling gene features, these alternatives did not outperform our adopted strategy (see below Table~\ref{table:gene}).

\begin{table}[t]
\caption{Simulated unimodal contributions in C-index (\%).}
\label{table:weight}
\centering
\resizebox{\linewidth}{!}{
\begin{tabular}{c|lllll|c}
    \toprule
    Modality Weight& BLCA & BRCA & CO-READ & HNSC & STAD &Mean\\
    \cmidrule(lr){1-7}
     $w_p=0.95,w_g=0.05$ & \fontsize{14}{16}\selectfont 63.0 \fontsize{10}{12}\selectfont$\pm$ 4.6 & \fontsize{14}{16}\selectfont 67.4 \fontsize{10}{12}\selectfont$\pm$ 10.8 & \fontsize{14}{16}\selectfont 70.3 \fontsize{10}{12}\selectfont$\pm$ 13.6 & \fontsize{14}{16}\selectfont 60.5 \fontsize{10}{12}\selectfont$\pm$ 5.1 & \fontsize{14}{16}\selectfont 63.7 \fontsize{10}{12}\selectfont$\pm$ 4.6  & \fontsize{14}{16}\selectfont65.0\\

     $w_p=0.05,w_g=0.95$ & \fontsize{14}{16}\selectfont 59.4 \fontsize{10}{12}\selectfont$\pm$ 4.9 & \fontsize{14}{16}\selectfont 61.7 \fontsize{10}{12}\selectfont$\pm$ 2.8 & \fontsize{14}{16}\selectfont 69.2 \fontsize{10}{12}\selectfont$\pm$ 14.1 & \fontsize{14}{16}\selectfont60.7 \fontsize{10}{12}\selectfont$\pm$ 4.7 & \fontsize{14}{16}\selectfont64.5 \fontsize{10}{12}\selectfont$\pm$ 4.5  & \fontsize{14}{16}\selectfont63.1\\
     
    \cmidrule(lr){1-7}
    Dynamic weighting  & \fontsize{14}{16}\selectfont \textbf{70.5} \fontsize{10}{12}\selectfont$\pm$ 4.1 & \fontsize{14}{16}\selectfont \textbf{72.9} \fontsize{10}{12}\selectfont$\pm$ 1.9 & \fontsize{14}{16}\selectfont\textbf{80.8} \fontsize{10}{12}\selectfont$\pm$ 5.8 & \fontsize{14}{16}\selectfont \textbf{66.0} \fontsize{10}{12}\selectfont$\pm$ 5.8 & \fontsize{14}{16}\selectfont \textbf{67.5} \fontsize{10}{12}\selectfont$\pm$ 3.3 & \fontsize{14}{16}\selectfont\textbf{71.5}\\

    \bottomrule
\end{tabular}
}

\end{table}

\subsection{Visualization for modality rebalance}
The rebalanced modality weights are shown in Fig.~\ref{fig:heatmap}.a (here). Before rebalancing, the dominance of high-dimensional pathology features suppressed the genomic contribution. After adjustment, the model places more emphasis on genomic features, mitigating the pathology-genomics imbalance. In comparison to Fig.~1 (main paper), which illustrates the original imbalance, Fig.~\ref{fig:heatmap}.b (here) demonstrates the effectiveness of our weighting strategy, seeing the shift toward lighter colors.

\subsection{Visualization for WSI and gene interactions}
Based on the BRCA visualization analysis in the main text, we also performed visualizations on other datasets. Visualization results for BLCA, CO-READ, HNSC, and STAD dataset are shown in Fig.~\ref{fig:blca}.

In the BLCA dataset, the elevated expression gradients of genes such as PAX7, LIMK2, UCN2, and RHOXF1 in low-risk patients suggest their involvement in cellular processes that may help reduce cancer aggressiveness. PAX7 is known for its role in muscle tissue development and repair, contributing to cellular stability and resilience. LIMK2 is involved in cytoskeletal organization, which might help maintain normal cell architecture and inhibit metastasis. UCN2, a member of the corticotropin-releasing factor family, could play a role in stress response modulation, which may help reduce tumor-promoting inflammation. RHOXF1 is implicated in transcriptional regulation, supporting pathways that suppress oncogenic activity.
Conversely, high-risk patients show higher gradients of MUTYH, SGK2, ERBB2L3, and MBTL1. MUTYH is involved in DNA repair; its dysregulation lead to genomic instability and increased mutation rates. SGK2 plays a role in cell survival and proliferation pathways, enhancing tumor growth. ERBB2L3 is related to the EGFR family of receptors, which are often associated with aggressive cancer phenotypes and treatment resistance. MBTL1 is involved in chromatin remodeling and gene expression regulation, which contribute to oncogenic transformation.

\begin{table}[t]
\caption{Genomic modeling in C-index (\%). `/' denotes our strategy.}

\label{table:gene}
\centering
\resizebox{\linewidth}{!}{

\begin{tabular}{c|lllll|c}
    \toprule
    Setting & BLCA & BRCA & CO-READ & HNSC & STAD &Mean\\
    \cmidrule(lr){1-7}
     GCN & \fontsize{14}{16}\selectfont 65.6 \fontsize{10}{12}\selectfont$\pm$ 5.6 & \fontsize{14}{16}\selectfont 70.2 \fontsize{10}{12}\selectfont$\pm$ 6.6 & \fontsize{14}{16}\selectfont 76.5 \fontsize{10}{12}\selectfont$\pm$ 9.0 & \fontsize{14}{16}\selectfont 62.5 \fontsize{10}{12}\selectfont$\pm$ 5.5 & \fontsize{14}{16}\selectfont 61.3 \fontsize{10}{12}\selectfont$\pm$ 4.5  & \fontsize{14}{16}\selectfont67.2\\

     GAT & \fontsize{14}{16}\selectfont 64.9 \fontsize{10}{12}\selectfont$\pm$ 2.5 & \fontsize{14}{16}\selectfont67.9 \fontsize{10}{12}\selectfont$\pm$ 5.8 & \fontsize{14}{16}\selectfont78.3 \fontsize{10}{12}\selectfont$\pm$ 12.0 & \fontsize{14}{16}\selectfont61.8 \fontsize{10}{12}\selectfont$\pm$ 6.0 & \fontsize{14}{16}\selectfont62.7 \fontsize{10}{12}\selectfont$\pm$ 4.8  & \fontsize{14}{16}\selectfont68.1 
     \\
     
    \cmidrule(lr){1-7}
    /  & \fontsize{14}{16}\selectfont \textbf{70.5} \fontsize{10}{12}\selectfont$\pm$ 4.1 & \fontsize{14}{16}\selectfont \textbf{72.9} \fontsize{10}{12}\selectfont$\pm$ 1.9 & \fontsize{14}{16}\selectfont\textbf{80.8} \fontsize{10}{12}\selectfont$\pm$ 5.8 & \fontsize{14}{16}\selectfont \textbf{66.0} \fontsize{10}{12}\selectfont$\pm$ 5.8 & \fontsize{14}{16}\selectfont \textbf{67.5} \fontsize{10}{12}\selectfont$\pm$ 3.3 & \fontsize{14}{16}\selectfont\textbf{71.5}\\

    \bottomrule
\end{tabular}
}
\end{table}

In the CO-READ dataset, low-risk patients exhibit high gradients of TSSK2, IL3RA, GATA4, and RETN. TSSK2 is involved in cellular signaling that contribute to tumor suppression. IL3RA plays a role in modulating immune responses to prevent cancer progression. GATA4 is a transcription factor important for cell differentiation that support pathways that maintain normal cellular function. RETN is linked to metabolic regulation and inflammation; its balanced expression might protect against tumor-promoting inflammatory processes.
High-risk patients display elevated gradients of GYPB, CAMK2B, DMRT1, and GIP. The involvement of GYPB in the structure of the erythrocyte membrane could reflect changes in the cellular environments conducive to tumor growth. CAMK2B's role in calcium signaling could enhance pathways that promote cancer cell survival and proliferation. The high gradient of DMRT1 could influence tumor behavior. GIP is involved in metabolic processes and contribute to creating an environment that supports cancer metabolism. The low gradient of IL2, crucial for T-cell proliferation, indicates impaired immune surveillance in high-risk patients, allowing for unchecked tumor growth.

In the HNSC dataset, low-risk patients show high expression gradients for genes such as GYPB, NEUROG3, and CALCB, while ACSL6 and EPHA5 exhibit low gradients. GYPB is involved in erythrocyte membrane structure, NEUROG3 plays a role in neurogenesis, and CALCB is linked to vasodilation and pain modulation. These genes contribute to protective mechanisms or less aggressive tumor behavior in low-risk patients.
In contrast, high-risk patients have elevated gradients of CALCB, DMRT1, and TNFRSF17, with IL2 and EPHA5 showing low gradients. The repeated high gradient of CALCB suggests its involvement in more complex roles depending on the risk category. DMRT1 is associated with sex differentiation, while TNFRSF17 is involved in B-cell development and indicate immune evasion tactics by the tumor. The low gradient of IL2, crucial for T-cell proliferation, might suggest impaired immune responses in high-risk patients. Similarly, reduced EPHA5 expression affect cell signaling pathways that normally suppress tumor growth.

In the STAD dataset, low-risk patients show high expression gradients for CAMK2B, DMRT1, and VIP, with IL2 exhibiting a negative high gradient. CAMK2B is involved in calcium signaling, contributing to cellular stability; DMRT1 plays a role in sex differentiation, which influence tumor suppression pathways; and VIP is associated with anti-inflammatory effects, reducing tumor progression. The negative gradient of IL2, important for T-cell proliferation, indicates a balanced immune response that prevents excessive inflammation.
In high-risk patients, PAX3, GYPB, and TSSK2 have high gradients, while IL2 also shows a negative high gradient. PAX3 is linked to cell growth and differentiation, which could drive tumor aggressiveness; GYPB is involved in erythrocyte function and reflect altered cellular environments; and TSSK2 is associated with cell signaling pathways that enhance tumor proliferation. The consistently negative gradient of IL2 suggests compromised immune responses in both risk groups and is more detrimental in high-risk patients due to the presence of other aggressive gene expressions.

\end{document}